\RequirePackage{fix-cm}
\documentclass[twocolumn,natbib]{svjour3}          % twocolumn
\smartqed  % flush right qed marks, e.g. at end of proof
\usepackage{graphicx}
\usepackage{mathptmx}      % use Times fonts if available on your TeX system
\usepackage{color,xcolor}
\usepackage{epsfig}
\usepackage{graphicx}

\usepackage{adjustbox}
\usepackage{array}
\usepackage{booktabs}
\usepackage{colortbl}
\usepackage{float,wrapfig}
\usepackage{hhline}
\usepackage{multirow}
\usepackage{subcaption} % issues a warning with CVPR/ICCV format

\usepackage{amsfonts}
\usepackage{mathtools}
\usepackage{bm}
\usepackage{nicefrac}

\usepackage{changepage}
\usepackage{extramarks}
\usepackage{fancyhdr}
\usepackage{lastpage}
\usepackage{setspace}
\usepackage{soul}
\usepackage{xspace}

\usepackage[pagebackref=true,breaklinks=true,colorlinks,citecolor=blue,bookmarks=false]{hyperref}
\usepackage{url}

\usepackage{algorithm, algorithmic}
\usepackage{enumerate}

\newcolumntype{L}[1]{>{\raggedright\let\newline\\\arraybackslash\hspace{0pt}}m{#1}}
\newcolumntype{C}[1]{>{\centering\let\newline\\\arraybackslash\hspace{0pt}}m{#1}}
\newcolumntype{R}[1]{>{\raggedleft\let\newline\\\arraybackslash\hspace{0pt}}m{#1}}

\newcommand{\xpar}[1]{\noindent\textbf{#1}\ \ }
\newcommand{\vpar}[1]{\vspace{3mm}\noindent\textbf{#1}\ \ }

\def\y{\mathit{y}}
\def\cX{\mathbf{X}}
\def\cY{\mathbf{Y}}

\def\bR{\mathbb{R}}

\newcommand{\tb}[1]{\textbf{#1}}

\newcommand{\sect}[1]{Section~\ref{#1}}

\newcommand{\eqn}[1]{Equation~\ref{#1}}
\newcommand{\fig}[1]{Figure~\ref{#1}}
\newcommand{\tbl}[1]{Table~\ref{#1}}

\newcommand{\ignorethis}[1]{}

\makeatletter
\DeclareRobustCommand\onedot{\futurelet\@let@token\@onedot}
\def\@onedot{\ifx\@let@token.\else.\null\fi\xspace}

\def\eg{\emph{e.g}\onedot}

\def\etc{\emph{etc}\onedot}

\makeatother

\definecolor{MyDarkBlue}{rgb}{0,0.08,1}
\definecolor{MyDarkGreen}{rgb}{0.02,0.6,0.02}
\definecolor{MyDarkRed}{rgb}{0.8,0.02,0.02}
\definecolor{MyDarkOrange}{rgb}{0.40,0.2,0.02}
\definecolor{MyPurple}{RGB}{111,0,255}
\definecolor{MyRed}{rgb}{1.0,0.0,0.0}
\definecolor{MyGold}{rgb}{0.75,0.6,0.12}
\definecolor{MyDarkgray}{rgb}{0.66, 0.66, 0.66}

\def\datasetName{Keypoint-5\xspace}
\def\model{3D INterpreter Networks\xspace}
\def\modelshort{3D-INN\xspace}

\journalname{International Journal of Computer Vision}
\begin{document}

\title{3D Interpreter Networks for Viewer-Centered Wireframe Modeling} 
\titlerunning{International Journal of Computer Vision}

\author{Jiajun Wu$^{1}$ \and
        Tianfan Xue$^{2}$ \and
        Joseph J. Lim$^3$ \and
        Yuandong Tian$^4$ \and \\
        Joshua B. Tenenbaum$^1$ \and
        Antonio Torralba$^1$ \and
        William T. Freeman$^{1,5}$
}

\authorrunning{International Journal of Computer Vision}

\institute{Jiajun Wu \at
           \email{jiajunwu@mit.edu}
           \and
           Tianfan Xue \at
           \email{tianfan@google.com}
           \and
           Joseph J. Lim \at
           \email{limjj@usc.edu}
           \and
           Yuandong Tian \at
           \email{yuandong@fb.com}
           \and
           Joshua B. Tenenbaum \at
           \email{jbt@mit.edu}
           \and
           Antonio Torralba \at
           \email{torralba@csail.mit.edu}
           \and
           William T. Freeman \at
           \email{billf@mit.edu}
           \and
           $^1$\;\; Massachusetts Institute of Technology, MA, USA \\
           $^2$\;\; Google Research, Mountain View, CA, USA \\
           $^3$\;\; University of Southern California, CA, USA \\
           $^4$\;\; Facebook Inc., CA, USA \\
           $^5$\;\; Google Research, Cambridge, MA, USA \\
           Jiajun Wu and Tianfan Xue contributed equally to this work. 
}

\date{Received: date / Accepted: date}

\maketitle

\begin{abstract}

Understanding 3D object structure from a single image is an important but challenging task in computer vision, mostly due to the lack of 3D object annotations to real images. Previous research tackled this problem by either searching for a 3D shape that best explains 2D annotations, or training purely on synthetic data with ground truth 3D information.

In this work, we propose \model (\modelshort), an end-to-end trainable framework that sequentially estimates 2D keypoint heatmaps and 3D object skeletons and poses. Our system learns from both 2D-annotated real images and synthetic 3D data. This is made possible mainly by two technical innovations. First, heatmaps of 2D keypoints serve as an intermediate representation to connect real and synthetic data. \modelshort is trained on real images to estimate 2D keypoint heatmaps from an input image; it then predicts 3D object structure from heatmaps using knowledge learned from synthetic 3D shapes. By doing so, \modelshort benefits from the variation and abundance of synthetic 3D objects, without suffering from the domain difference between real and synthesized images, often due to imperfect rendering. Second, we propose a Projection Layer, mapping estimated 3D structure back to 2D. During training, it ensures \modelshort to predict 3D structure whose projection is consistent with the 2D annotations to real images. 

Experiments show that the proposed system performs well on both 2D keypoint estimation and 3D structure recovery. We also demonstrate that the recovered 3D information has wide vision applications, such as image retrieval.


\end{abstract}

\section{Introduction}
\label{sec:intro}

\begin{figure*}[t]
	\centering
	\includegraphics[width=\linewidth]{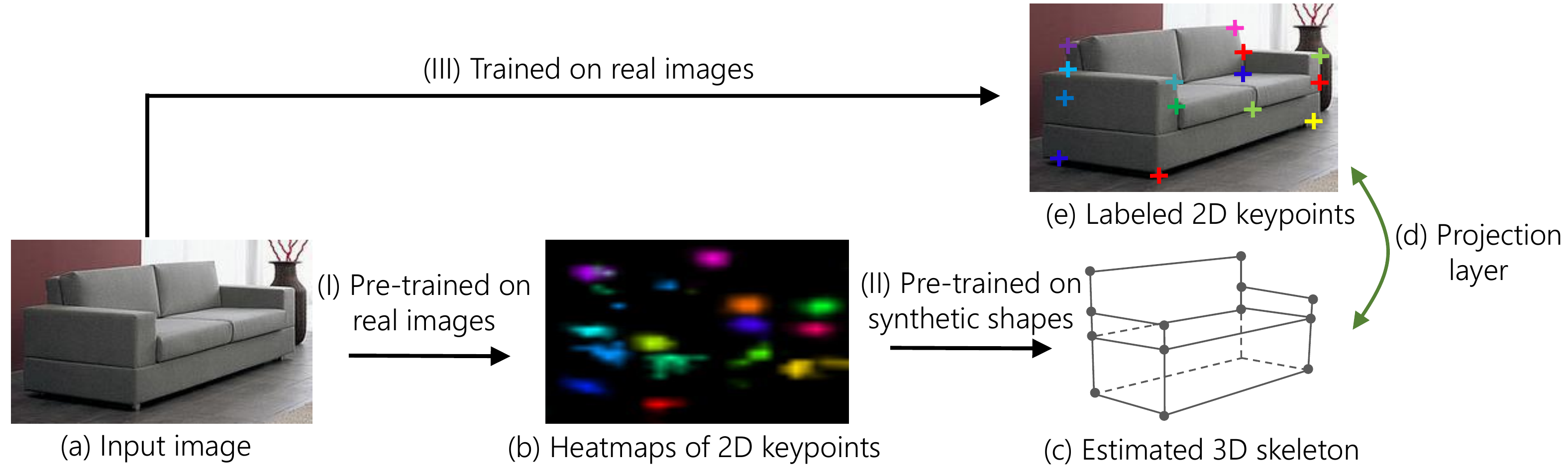}
	\caption{An overview of our model. For an image (a) with a category-level label (sofa), the system first estimates its 2D keypoint heatmaps (b), and then recovers the 3D skeleton of the object (c). During training, through the projection layer (d), it also enforces the consistency between annotated 2D keypoint locations (e) and projected 2D locations of estimated 3D keypoints.}
	\label{fig:overview}
\end{figure*}

\begin{figure}[t]
	\centering
	\includegraphics[width=\linewidth]{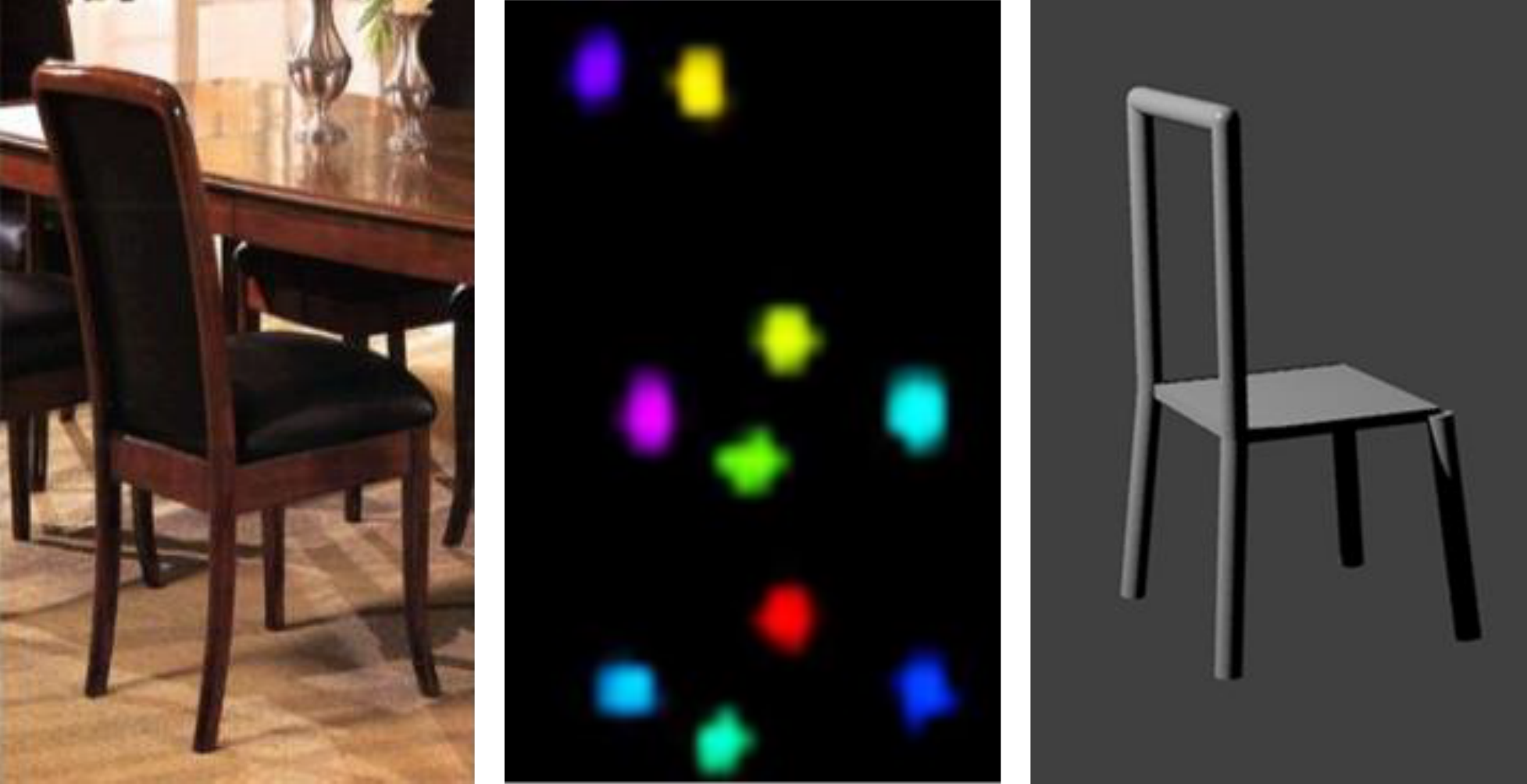}
	\caption{Given an image of a chair, we are interested in its intrinsic properties such as its height, leg length, and seat width, and extrinsic properties such as its pose.}
	\label{fig:teaser}
\end{figure}

Deep networks have achieved impressive performance on image recognition~\citep{russakovsky2015imagenet}. Nonetheless, for any visual system to parse objects in the real world, it needs not only to assign category labels to objects, but also to interpret their intra-class variation. \fig{fig:teaser} shows an example: for a chair, we are interested in its intrinsic properties such as its \emph{style}, \emph{height}, leg \emph{length}, and seat \emph{width}, and extrinsic properties such as its \emph{pose}.

In this paper, we recover these object properties from a single image by jointly estimating the object's 3D wireframe and the viewpoint. We choose to use a single image as input primarily for three reasons. First, this problem has great scientific value: humans can easily recognize 3D structure from a single image, and we want to build machines that replicate such an ability. Second, by starting with a single image, instead of multi-view images or videos, our model can be directly applied to images taken in the wild, \eg, from the web, and can cluster them based on their structure or pose. This offers extensive practical usages: social media and e-commerce companies can better understand and analyze user-uploaded data, and household robots can efficiently interact with objects after modeling them in 3D. Third, using a single image as input enables online inference: for moving objects like cars, the system can reconstruct their geometry and viewpoint on the fly. This is crucial for real-time applications such as autonomous driving.

We represent an object via a 3D skeleton~\citep{non_rigid3d} (\fig{fig:overview}c), instead of a 3D mesh or a depth map~\citep{DB15,Aubry14,prasad2010finding,kar2015category,haosu_sig14,vicente2014reconstructing,huang2015single}, because skeletons are simpler and preserve the structural properties that we are interested in. We refer readers to \sect{sec:discuss} for detailed discussions on our design choices. We assume that a 3D skeleton consists of keypoints and the connections between them, and manually pre-define the skeleton model for each object category (\eg, chair, sofa, and car). Then, our task is to estimate 3D keypoint locations of an object from a single RGB image. 

The main challenge of single image 3D estimation is the difficulty in obtaining training images with ground truth 3D structure, as manually annotating 3D object structure in real images is labor-intensive and often inaccurate. Previous methods tackled this problem mostly in two ways. One is to directly solve for a 3D skeleton from estimated 2D keypoint locations by minimizing its reprojection error~\citep{zhou153d}, without any 3D annotations. Most algorithms in this category do not encode 3D shape priors within the model, and thus they are not robust to inaccurate keypoint estimation, as shown in experiments in \sect{sec:results}. The other is to train on synthetically rendered images of 3D objects~\citep{li2015joint,su15}; in synthetic data, complete 3D structure is available. But the statistics of synthesized images are often different from those of real images, due to changes in lighting, occlusion, and shape details. This makes it hard for models trained only on synthetic data to generalize well to real images. 

In this paper, we propose \model (\modelshort), an end-to-end trainable framework for 3D skeleton and viewpoint estimation. In contrast to prior art, our model learns from both 2D-labeled real images and synthetic 3D objects. This is made possible by two major innovations. First, we use heatmaps of 2D keypoints as an intermediate representation to connect real and synthetic data. \modelshort is trained on real images to estimate 2D keypoint heatmaps from an input image (\fig{fig:overview}-I); it then learns from synthetic data to estimate 3D structure from heatmaps (\fig{fig:overview}-II). By doing so, \modelshort benefits from the variations in abundant synthetic 3D objects, without suffering from the domain difference between real and synthesized data.

Second, we introduce a \emph{Projection Layer}, a rendering function that calculates projected 2D keypoint positions given a 3D skeleton and camera parameters. We attach it at the end of the framework (\fig{fig:overview}d). This enables \modelshort to predict 3D structural parameters that minimize errors in 2D on real images with keypoint labels, without requiring 3D object annotations. Our training paradigm therefore consists of three steps: we first train the keypoint estimation component (\fig{fig:overview}-I) on 2D-annotated real images; we then train the 3D interpreter (\fig{fig:overview}-II) on synthetic 3D data; we finally fine-tune the entire framework end-to-end with the projection layer (\fig{fig:overview}-III).

Both innovations are essential for the system to exploit the complementary richness of real 2D data and synthetic 3D data. Without the heatmap representation and synthetic 3D data, the system can still be trained on real images with 2D annotations, using the projection layer. But it does not perform well: because of intrinsic ambiguities in 2D-to-3D mapping, the algorithm recovers unnatural 3D geometries, though their projections may perfectly align with 2D annotations, as explored in~\cite{lowe1987three}. Synthetic 3D shapes help the network resolve this ambiguity by learning prior knowledge of ``plausible shapes''. Without the projection layer, the system becomes two separately trained networks: one trained on real images to estimate 2D keypoint heatmaps, and the other trained on synthetic data to estimate 3D structure. As shown in \sect{sec:results}, the 3D predictions in this case are not as accurate due to the domain adaptation issue.

Several experiments demonstrate the effectiveness of \modelshort. First, the proposed network achieves good performance on various keypoint localization datasets, including FLIC \citep{FLIC} for human bodies, CUB-200-2011 \citep{CUB} for birds, and our new dataset, \datasetName, for furniture. We then evaluate our network on IKEA~\citep{ikea}, a dataset with ground truth 3D object structure and viewpoints. We augmented the original IKEA dataset with additional 2D keypoint labels. On 3D structure estimation, \modelshort shows its advantage over an optimization-based method \citep{zhou153d} when keypoint estimation is imperfect. On 3D viewpoint estimation, it also performs better than the state of the art~\citep{su15}. We further evaluate \modelshort, in combination with an object detection framework, R-CNN~\citep{RCNN}, on PASCAL 3D+ benchmark~\citep{pascal3d} for joint detection and viewpoint estimation. \modelshort also achieves results comparable to the state of the art~\citep{su15,tulsiani2015viewpoints}. At last, we show that \modelshort has wide vision applications including 3D object retrieval.

Our contributions are three-fold. First, we introduce \model (\modelshort); by incorporating 2D keypoint heatmaps to connect real and synthetic worlds, we strengthen the generalization ability of the network. Second, we propose a projection layer, so that \modelshort can be trained to predict 3D structural parameters using only 2D-annotated images. Third, our model achieves state-of-the-art performance on both 2D keypoint and 3D structure and viewpoint estimation.

\section{Related work}

\xpar{Single Image 3D Reconstruction} Previous 3D reconstruction methods mainly modeled objects using either dense representations such as depth or meshes, or sparse representations such as skeletons or pictorial structure. Depth-/mesh-based models can recover detailed 3D object structure from a single image, either by adapting existing 3D models from a database~\citep{Aubry14,satkin_bmvc2012,haosu_sig14,huang2015single,zeng20163dmatch,3dgan,hu2015learning,bansal2016marr,shrivastava2013building,choy20163d}, or by inferring from its detected 2D silhouette~\citep{kar2015category,Soltani2017Synthesizing,vicente2014reconstructing,prasad2010finding,marrnet}. 

In this paper, we choose to use a skeleton-based representation, exploiting the power of abstraction. The skeleton model can capture geometric changes of articulated objects~\citep{non_rigid3d,yasin2016dualsource,akhter2015pose}, like a human body or the base of a swivel chair. Typically, researchers recovered a 3D skeleton from a single image by minimizing its projection error on the 2D image plane~\citep{lowe1987three,leclerc1992optimization,synthesis3d,xue2012example,ramakrishna2012reconstructing,zia2013detailed}. Recent work in this line \citep{akhter2015pose,zhou153d} demonstrated state-of-the-art performance. In contrast to them, we propose to use neural networks to predict a 3D object skeleton from its 2D keypoints, which is more robust to imperfect detection results and can be jointly learned with keypoint estimators.

Our work also connects to the traditional field of vision as inverse graphics~\citep{hinton1997generative,kulkarni2015deep} and analysis by synthesis~\citep{yuille2006vision,kulkarni2015picture,bever2010analysis,wu2015galileo}, as we use neural nets to decode latent 3D structure from images, and use a projection layer for rendering. Their approaches often required supervision for the inferred representations or made over-simplified assumptions of background and occlusion in images. Our \modelshort learns 3D representation without using 3D supervision, and generalizes to real images well.

\vpar{2D Keypoint Estimation} 
Another line of related work is 2D keypoint estimation. During the past decade, researchers have made significant progress in estimating keypoints on humans~\citep{FLIC,yang2011articulated} and other objects~\citep{CUB,shih2015part}. Recently, there have been several attempts to apply convolutional neural networks to human keypoint estimation~\citep{toshev2014deeppose,tompson2015efficient,carreira2015human,newell2016stacked}, which all achieved significant improvement. \modelshort uses 2D keypoints as an intermediate representation, and aims to recover a 3D skeleton from them.

\vpar{3D Viewpoint Estimation} 
3D viewpoint estimation seeks to estimate the 3D orientation of an object from a single image~\citep{pascal3d}. Some previous methods formulated it as a classification or regression problem, and aimed to directly estimate the viewpoint from an image~\citep{urtasun_3d_car,su15}. Others proposed to estimate 3D viewpoint from detected 2D keypoints or edges in the image~\citep{zia2013detailed,fpm,tulsiani2015viewpoints}. While the main focus of our work is to estimate 3D object structure, our method can also predict the corresponding 3D viewpoint. 

\vpar{Training with Synthetic Data} 
Synthetic data are often used to augment the training set~\citep{haosu_sig14,shakhnarovich2003fast}, especially when ground truth labels of real images are hard to obtain. This technique has found wide applications in computer vision. To name a few, \cite{sun2014virtual} and \cite{zhou2016learning} combined real and synthetic data for object detection and matching, respectively. \cite{huang2015single} analyzed the invariance of convolutional neural networks using synthetic images. \cite{DB15} trained a neural network for image synthesis using synthetic images. \cite{mccormac2017scenenet} rendered images for indoor scene understanding. \cite{haosu_sig14} attempted to train a 3D viewpoint estimator on both real and synthetic images.

In this paper, we combine real 2D-annotated images and synthetic 3D data for training \modelshort to recover a 3D skeleton. We use heatmaps of 2D keypoints, instead of (often imperfectly) rendered images, from synthetic 3D data, so that our algorithm has better generalization ability as the effects of imperfect rendering are minimized. \cite{yasin2016dualsource} also proposed to use both 2D and 3D data for training, but they used keypoint locations, instead of heatmaps, as the intermediate representation that connects 2D and 3D. While their focus is on estimating human poses, we study the problem of recovering the 3D structure of furniture and cars.

\begin{figure*}[t]
	\centering
    
    \begin{subfigure}{\linewidth}
    \centering
    \includegraphics[width=0.5\linewidth]{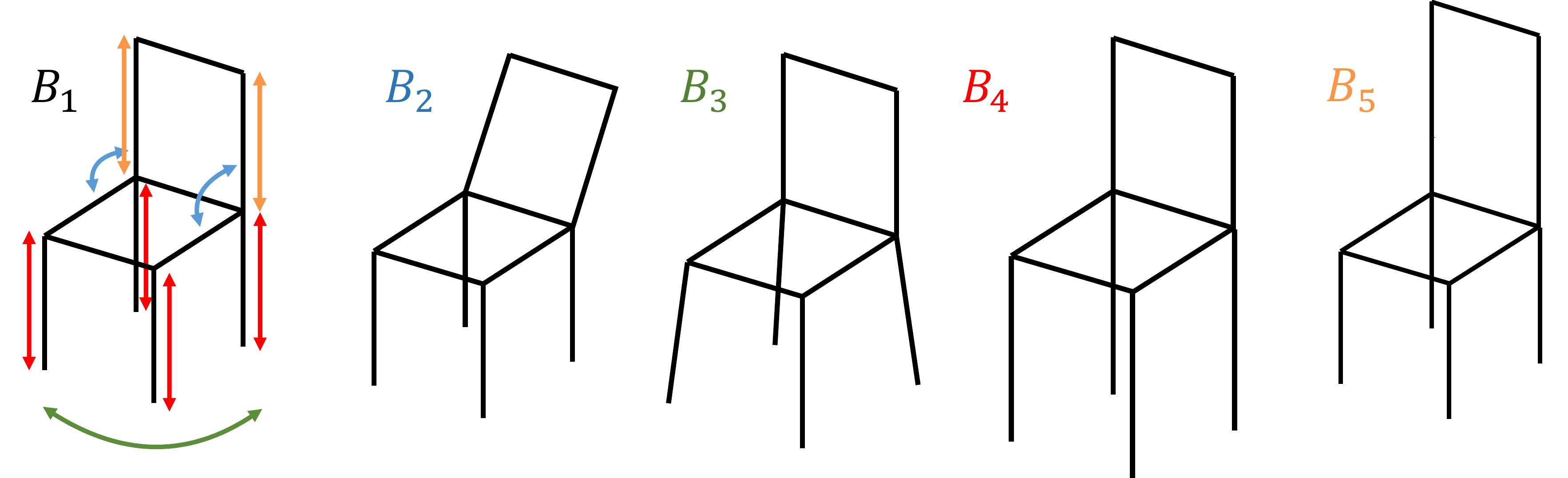}
    \caption{Base shapes for chairs. $B_1$ is the mean shape of chairs, and the others characterize possible variations of the structure.}
    \label{fig:syn_chair}
    \end{subfigure}
    
    \vspace{15pt}
	\begin{subfigure}{\linewidth}
    \centering
    \includegraphics[width=\linewidth]{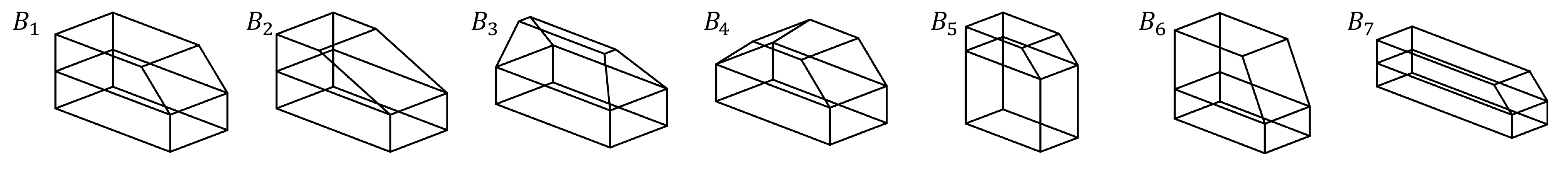}
    \caption{Base shapes for cars}
    \label{fig:syn_car}
    \end{subfigure}
	
	\caption{Our skeleton model and base shapes for chairs (a) and cars (b).}
	\label{fig:syn}
\end{figure*}

\section{Method}
\label{sec:methods}

\begin{figure*}[t]
	\centering
	\includegraphics[width=\linewidth]{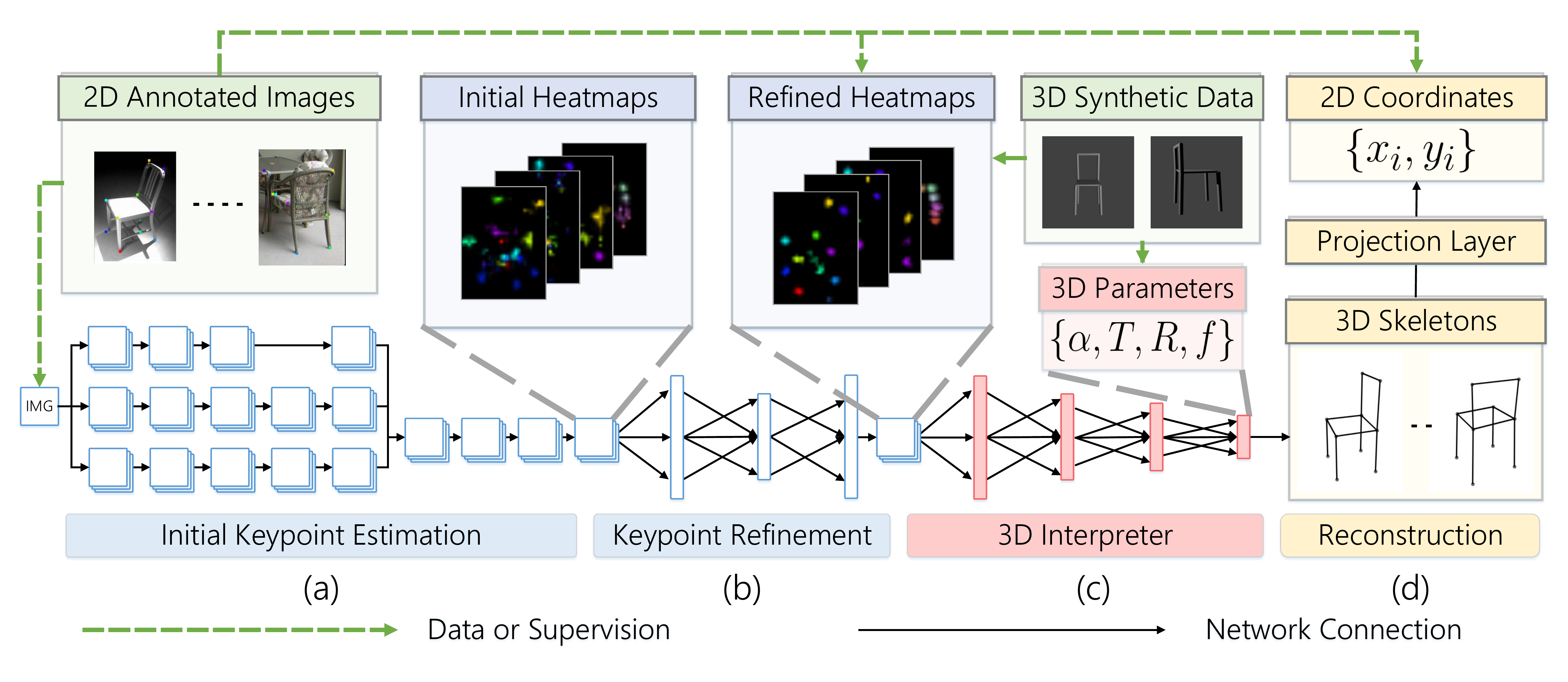}
	\caption{\modelshort takes a single image as input and reconstructs the detailed 3D structure of the object in the image (\eg, human, chair, \etc). The network is trained independently for each category, and here we use chairs as an example. \textbf{(a)} Estimating 2D keypoint heatmaps with a multi-scale CNN. \textbf{(b)} Refining keypoint locations by considering the structural constraints between keypoints. This is implicitly enforced with an information bottleneck and yields cleaner heatmaps. \textbf{(c)} Recovered 3D structural and camera parameters $\{\alpha, T, R, f\}$. \textbf{(d)} The projection layer maps reconstructed 3D skeletons back to 2D keypoint coordinates.} 
	\label{fig:architecture} 
\end{figure*}

We design a deep convolutional network to recover 3D object structure. The input to the network is a single image with the object of interest at its center, which can be obtained by state-of-the-art object detectors. The output of the network is a 3D object skeleton, including its 2D keypoint locations, 3D structural parameters, and 3D pose (see \fig{fig:architecture}). In the following sections, we will describe our 3D skeleton representation and the camera model (\sect{sec:skeleton}), network design (\sect{sec:network}), and training strategy (\sect{sec:synthetic_data}).

\subsection{3D Skeleton Representation}
\label{sec:skeleton}

We use skeletons as our 3D object representation. A skeleton consists of a set of keypoints as well as their connections. For each object category, we manually design a 3D skeleton characterizing its abstract 3D geometry. 

There exist intrinsic ambiguities in recovering 3D keypoint locations from a single 2D image. We resolve this issue by assuming that objects can only have constrained deformations~\citep{non_rigid3d}. For example, chairs may have various leg lengths, but for a single chair, its four legs are typically of equal length. We model these constraints by formulating 3D keypoint locations as a weighted sum of a set of base shapes~\citep{kar2015category}. The first base shape is the mean shape of all objects within the category, and the rest define possible deformations and intra-class variations. \fig{fig:syn_chair} shows our skeleton representation for chairs: the first is the mean shape of chairs, the second controls how the back bends, and the last two are for legs. \fig{fig:syn_car} shows base shapes for cars. The weight for each base shape determines how strong the deformation is, and we denote these weights as the \emph{structural parameters} of an object.

Formally, let $\cY \in \bR^{3\times N}$ be a matrix of 3D coordinates of all $N$ keypoints. Our assumption is that the 3D keypoint locations are a weighted sum of base shapes $B_k\in\bR^{3\times N}$, or 
\begin{equation}
\cY = \sum_{k=1}^K \alpha_k B_k,
\end{equation}
where $\{\alpha_k\}$ is the set of structural parameters of this object, and $K$ is the number of base shapes. 

Further, let $\cX \in \bR^{2\times N}$ be the corresponding 2D coordinates. Then the relationship between the observed 2D coordinates $\cX$ and the structural parameters $\{\alpha_k\}$ is
\begin{equation}
\cX = P(R\cY + T) = P(R\sum_{k=1}^K \alpha_k B_k + T),
\label{eq:pnp}
\end{equation}
where $R\in\bR^{3\times 3}$ (rotation) and $T\in\bR^{3}$ (translation) are the external parameters of the camera, and $P\in\bR^{3\times 4}$ is the camera projection matrix which we will discuss soon.

Therefore, to recover the 3D structural information of an object in a 2D image, we only need to estimate its structural parameters ($\{\alpha_k\}$) and the external viewpoint parameters ($R$, $T$, and $f$). We supply the detailed camera model below. In \sect{sec:network} and \sect{sec:synthetic_data}, we discuss how we design a neural network for this task, and how it can be jointly trained with real 2D images and synthetic 3D objects.

\vpar{Camera Model} We use perspective projection in order to model the perspective distortion in 2D images. We assume that the principal point is at the origin, all pixels are square, and there is no axis skew. In this way, we only need to estimate the focal length in the camera projection matrix $P$.

For the ease of inference in neural networks, we rewrite the normal perspective projection as follows. Let $x_i \in \bR^{2}$ be a column vector of the 2D coordinates of the $i$-th keypoint and $\y_i$ be the corresponding 3D coordinates to be recovered. We assume that the camera center is at $(0,0,f)$ instead of the origin. The perspective projection is written as (using projective coordinates):
\begin{align}
\centering
\left( \begin{array}{c} x^1_i \\ x^2_i \\ 1 \end{array} \right) =
\left(\begin{array}{cccc} f & 0 & 0 & 0 \\ 0 & f & 0 & 0 \\ 0 & 0 & 1 & 0 \end{array} \right)
\left(\begin{array}{c} y^1_i \\ y^2_i \\ y^3_i + f \\ 1\end{array}  \right),
\label{eqn:proj}
\end{align}
where $f$ is the focal length, $x^1_i$ and $x^2_i$ are the x- and y-components of $x_i$, and $y^1_i$, $y^2_i$, and $y^3_i$ are x-, y-, and z-components of $y_i$. 

When $f^{-1}\rightarrow 0$, \eqn{eqn:proj} converges to the formulation of parallel projection. To see that, based on \eqn{eqn:proj}, we get the Euclidean coordinates of the 2D projection as (we abuse the notation of $x^1_i$ and $x^2_i$ for both Euclidean coordinates and projective coordinates)
\begin{align}
\centering
\begin{cases}
x^1_i = \dfrac{fy_i^1}{y^3 + f} = \dfrac{y_i^1}{f^{-1}y^3 + 1}, \\
x^2_i = \dfrac{fy_i^2}{y^3 + f} = \dfrac{y_i^2}{f^{-1}y^3 + 1}.
\end{cases}
\end{align}
Then when $f\rightarrow \infty$, we have
\begin{align}
\centering
\begin{cases}
x^1_i = \y_i^1, \\
x^2_i = \y_i^2,
\end{cases}
\end{align}
which is the formulation of parallel projection. Therefore, \eqn{eqn:proj} models the perspective projection when $f^{-1} \neq 0$ and models the parallel projection when $f^{-1} = 0$.

\subsection{\model(\modelshort)}
\label{sec:network}

Our network consists of three components: first, a keypoint estimator, which localizes 2D keypoints of objects from 2D images by regressing their heatmaps (\fig{fig:architecture}a and b, in blue); second, a 3D interpreter, which infers internal 3D structural and viewpoint parameters from the heatmaps (\fig{fig:architecture}c, in red); third, a projection layer, mapping 3D skeletons to 2D keypoint locations so that real 2D-annotated images can be used as supervision (\fig{fig:architecture}d, in yellow).

\vpar{Keypoint Estimation}
The keypoint estimation consists of two steps: initial estimation (\fig{fig:architecture}a) and keypoint refinement (\fig{fig:architecture}b). 

The network architecture for initial keypoint estimation is inspired by the pipeline proposed by \cite{tompson2014joint,tompson2015efficient}. The network takes multi-scaled images as input and estimates keypoint heatmaps. Specifically, we apply Local Contrast Normalization (LCN) on each image, and then scale it to 320$\times$240, 160$\times$120, and 80$\times$60 as input to three separate scales of the network. The output is $k$ heatmaps, each with resolution 40$\times$30, where $k$ is the number of keypoints of the object in the image. 

At each scale, the network has three sets of 5$\times$5 convolutional (with zero padding), ReLU, and 2$\times$2 pooling layers, followed by a 9$\times$9 convolutional and ReLU layer. The final outputs for the three scales are therefore images with resolution 40$\times$30, 20$\times$15, and 10$\times$7, respectively. We then upsample the outputs of the last two scales to ensure they have the same resolution (40$\times$30). The outputs from the three scales are later summed up and sent to a Batch Normalization layer and three 1$\times$1 convolution layers, whose goal is to regress target heatmaps. We found that Batch Normalization is critical for convergence, while Spatial Dropout, proposed in \cite{tompson2015efficient}, does not affect performance.

The second step of keypoint estimation is keypoint refinement, whose goal is to implicitly learn category-level structural constraints on keypoint locations after the initial keypoint localization. The motivation is to exploit the contextual and structural knowledge among keypoints (\eg, arms cannot be too far from the torso). We design a mini-network which, like an autoencoder, has information bottleneck layers, enforcing it to implicitly model the relationship among keypoints. Some previous works also use this idea and achieve better performance with lower computational cost in object detection~\citep{FastRCNN} and face recognition~\citep{deepface2}.

\begin{figure*}[t!]
    \centering
	\includegraphics[width=0.75\linewidth]{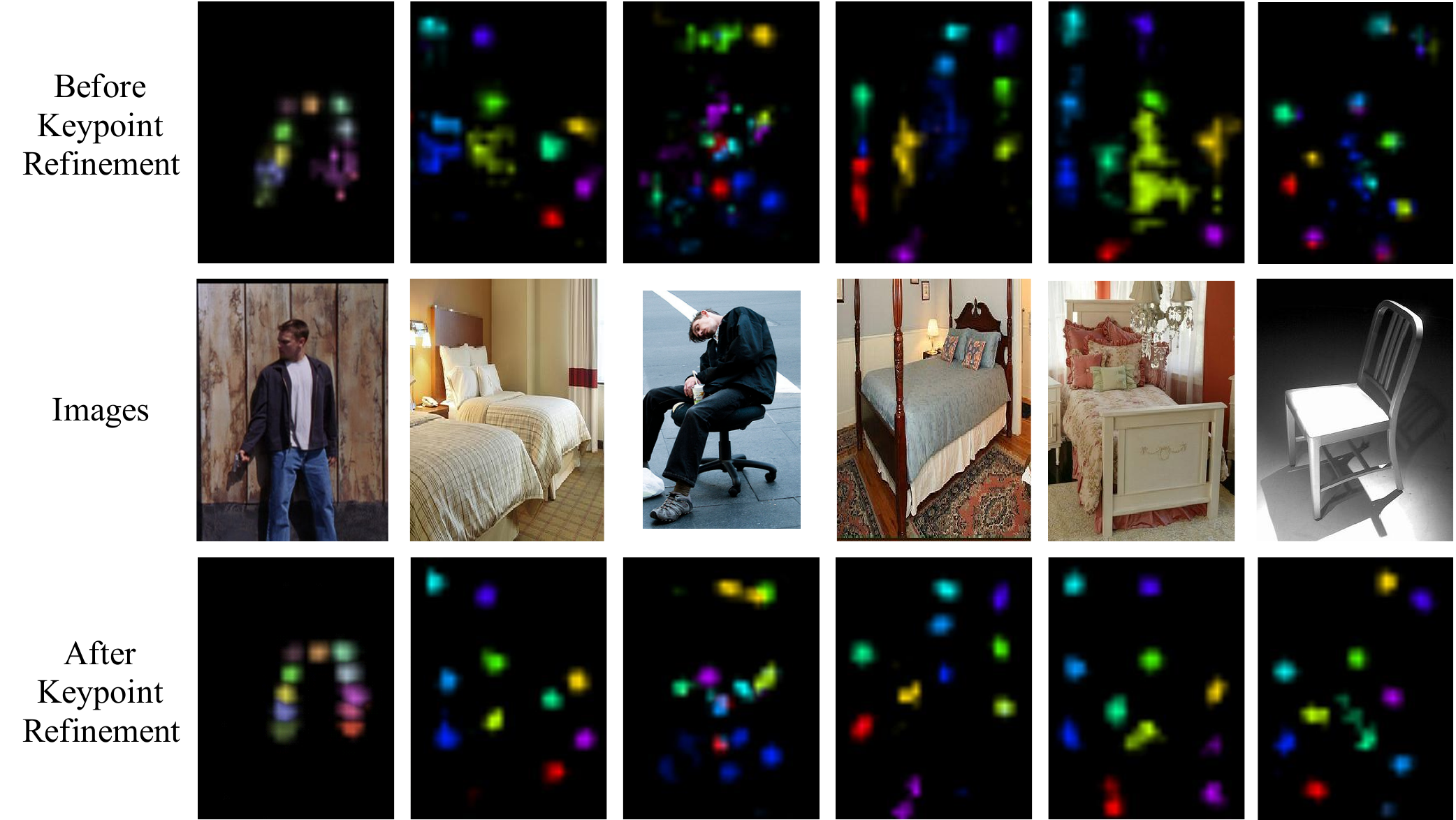}
    \caption{2D keypoint predictions from a single image, where each color corresponds to a keypoint. The keypoint refinement step cleans up false positives and produces more regulated predictions. }
    \label{fig:sl}
\end{figure*}

In the keypoint refinement network, we use three fully connected layers with widths 8,192, 4,096, and 8,192, respectively. After refinement, the heatmaps of keypoints are much cleaner, as shown in \fig{fig:sl} and \sect{sec:results}. 

\vpar{3D Interpreter}
The goal of our 3D interpreter is to infer 3D structure and viewpoint parameters, using estimated 2D heatmaps from earlier layers. While there are many different ways of solving \eqn{eq:pnp}, our deep learning approach has clear advantages. First, traditional methods~\citep{hejrati2012analyzing,non_rigid3d} that minimize the reprojection error consider only one keypoint hypothesis, and is therefore not robust to noisy keypoint detection. In contrast, our framework uses soft heatmaps of keypoint locations as input, as shown in \fig{fig:architecture}c, which is more robust when some keypoints are invisible or incorrectly located. Second, our algorithm only requires a single forward propagation during testing, making it more efficient than the most previous optimization-base methods.

As discussed in \sect{sec:skeleton}, the set of 3D parameters we estimate is of $S=\{\alpha_i, R, T, f\}$, with which we are able to recover the 3D object structure using \eqn{eq:pnp}. In our implementation, the network predicts $f^{-1}$ instead of $f$ for better numerical stability. As shown in \fig{fig:architecture}c, we use four fully connected layers as our 3D interpreter, with widths 2,048, 512, 128, and $|S|$, respectively. Spatial Transformer Networks~\citep{jaderberg2015spatial} also explored the idea of learning rotation parameters $R$ with neural nets, but our network can also recover structural parameters $\{\alpha_i\}$. 

\vpar{Projection Layer}
The last component of the network is a projection layer (\fig{fig:architecture}d). The projection layer takes estimated 3D parameters as input, and computes projected 2D keypoint coordinates $\{x_i,y_i\}$ using \eqn{eq:pnp}. As all operations are differentiable, the projection layer enables us to use 2D-annotated images as ground truth, and run back-propagation to update the entire network.

\subsection{Training Strategy}
\label{sec:synthetic_data}

A straightforward training strategy is to use real 2D images as input, and their 2D keypoint locations as supervision for the output of the projection layer. Unfortunately, experiments show that the network can hardly converge using this training scheme, due to the high-dimensional search space and the ambiguity in the 3D to 2D projection. 

We therefore adopt an alternative three-step training strategy: first, training the keypoint estimator (\fig{fig:architecture}a and \ref{fig:architecture}b) using real images with 2D keypoint heatmaps as supervision; second, training the 3D interpreter (\fig{fig:architecture}c) using synthetic 3D data as there are no ground truth 3D annotations available for real images; and third, training the whole network using real 2D images with supervision on the output of the projection layer at the end.

\begin{figure*}[t]
	\centering
	
	\begin{subfigure}{0.4\linewidth}
	\includegraphics[trim={0 0 27cm 0},clip,width=\linewidth]{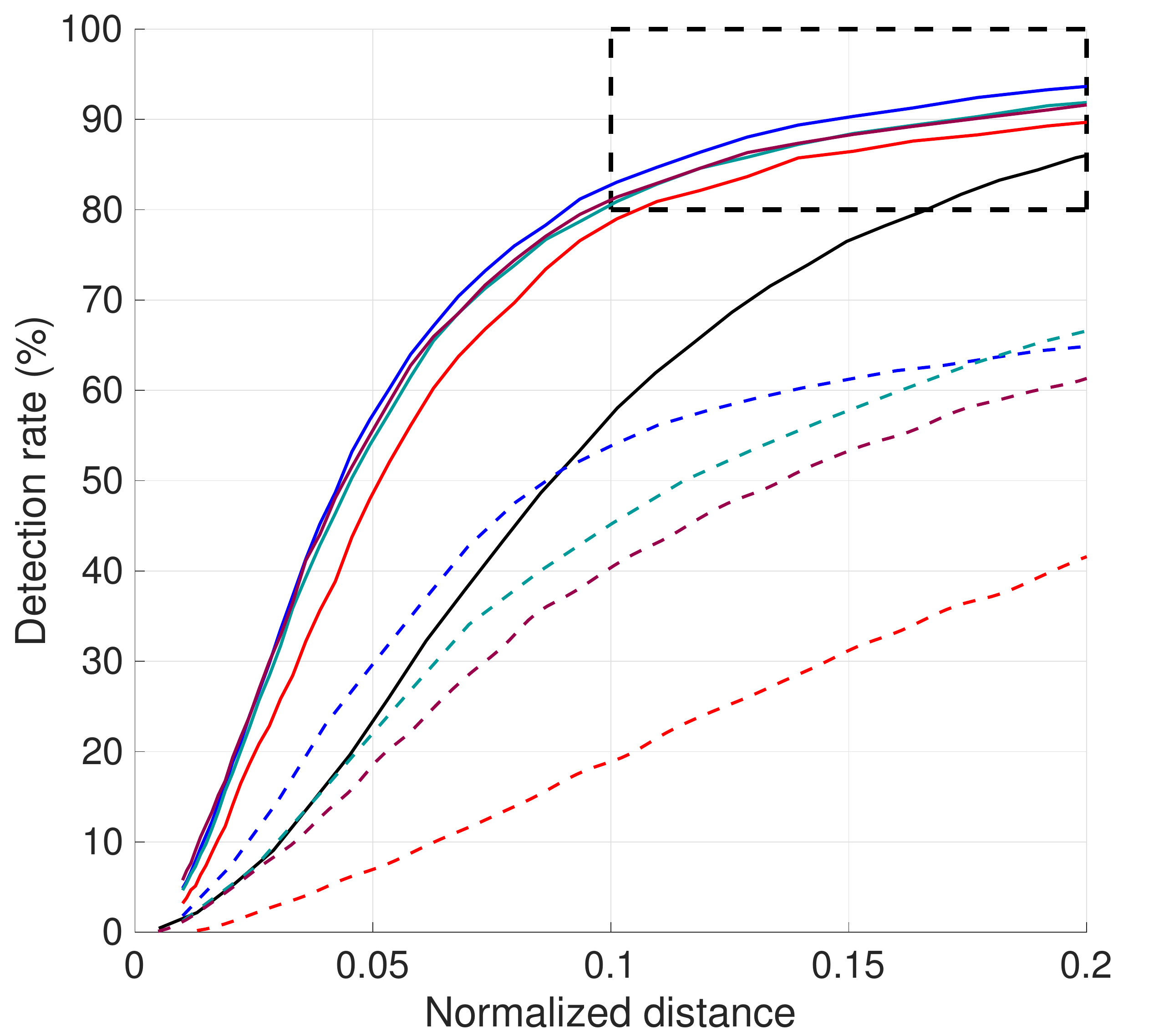}
	\caption{PCK curves on the FLIC dataset}
	\end{subfigure}
	\;\;
	\begin{subfigure}{0.4\linewidth}
	\includegraphics[trim={27cm 0 0 0},clip,width=\linewidth]{fig/flic_curve.pdf}
	\caption{A zoomed view of the dashed rectangle in (a)}
	\end{subfigure}
	
	\caption{\label{fig:human_keypoint_curves} (a) PCK curves on the FLIC dataset~\citep{FLIC}. \modelshort performs consistently better than other methods. Without keypoint refinement, it is comparable to \cite{tompson2015efficient}. (b) A zoomed view of the dashed rectangle in (a).}
\end{figure*}

To generate synthetic 3D objects, for each object category, we first randomly sample structural parameters $\{\alpha_i\}$ and viewpoint parameters $P$, $R$ and $T$. We calculate 3D keypoint coordinates using \eqn{eq:pnp}. To model deformations that cannot be captured by base shapes, we add Gaussian perturbation to 3D keypoint locations of each synthetic 3D object, whose variance is 1\% of its diagonal length.  Examples of synthetic 3D shapes are shown in \fig{fig:architecture}c. In experiments, we do not render synthesized shapes; we use heatmaps of keypoints, rather than rendered images, as training input.

\section{Evaluation} \label{sec:results}

We evaluate our entire framework, \modelshort, as well as each component within. In this section, we present both qualitative and quantitative results on 2D keypoint estimation (\sect{sec:eval_2d}) and 3D structure and viewpoint recovery (\sect{sec:eval_3d}).

\subsection{2D Keypoint Estimation}
\label{sec:eval_2d}

\xpar{Data}
For 2D keypoint estimation, we evaluate our algorithm on three image datasets: FLIC~\citep{FLIC} for human bodies, CUB-200-2011~\citep{CUB} for birds, and a new dataset \datasetName for furniture. Specifically, FLIC is a challenging dataset containing 3,987 training images and 1,016 test images, each labeled with 10 keypoints of human bodies. The CUB-200-2011 dataset was originally proposed for fine-grained bird classification, but with labeled keypoints of bird parts. It has 5,994 images for training and 5,794 images for testing, each coming with up to 15 keypoints.

We also introduce a new dataset, \datasetName, which contains five categories: bed, chair, sofa, swivel chair, and table. There are 1,000 to 2,000 images in each category, where 80\% are for training and 20\% for testing. For each image, we asked three workers on Amazon Mechanical Turk to label locations of a pre-defined category-specific set of keypoints; we then, for each keypoint, used the median of the three labeled locations as ground truth.

\vpar{Metrics}
To quantitatively evaluate the accuracy of estimated keypoints on FLIC (human body), we use the standard Percentage of Correct Keypoints (PCK) measure~\citep{FLIC} to be consistent with previous works~\citep{FLIC,tompson2014joint,tompson2015efficient}. We use the evaluation toolkit and results of competing methods released by  \cite{tompson2015efficient}. On CUB-200-2011 (bird) and the new \datasetName (furniture) dataset, following the convention~\citep{liu2013bird,shih2015part}, we evaluate results in Percentage of Correct Parts (PCP) and Average Error (AE). PCP is defined as the percentage of keypoints localized within 1.5 times of the standard deviation of annotations. We use the evaluation code from~\citep{liu2013bird} to ensure consistency. Average error is computed as the mean of the distance, bounded by 5, between a predicted keypoint location and ground truth. 

\begin{table}[t!]
    \centering
    \small
    \caption{\label{table:bird_keypoint_result}Keypoint estimation results on CUB-200-2011, measured in PCP (\%) and AE. Our method is comparable to Mdshift~\citep{shih2015part} (better in AE but worse in PCP), and better than all other algorithms.}
    \begin{tabular}{lcccc}
    \toprule
    Method & PCP (\%) & Average Error \\
    \midrule
    Poselets~\citep{poselet} & 27.47 & 2.89 \\
    Consensus~\citep{consensus} & 48.70 & 2.13 \\
    Exemplar~\citep{liu2013bird} & 59.74 & 1.80 \\
    Mdshift~\citep{shih2015part} & {\bf 69.1} & 1.39 \\
    \modelshort (ours) & 66.7 & {\bf 1.36} \\
    \midrule
    Human & 84.72 & 1.00 \\
    \bottomrule
    \end{tabular}
\end{table}

\begin{table}[t!]
    \centering
	\small
    \caption{\label{table:other_keypoint} Keypoint estimation results of \modelshort and \cite{tompson2015efficient} on \datasetName, measured in PCP (\%) and AE.  \modelshort is consistently better in both measures. We retrained the network in \cite{tompson2015efficient} on \datasetName.}
    \begin{tabular}{llcccc}
    \toprule
     & Method & Bed & Chair & Sofa & Swivel Chair \\
    \midrule
    \multirow{2}{*}{PCP} & \modelshort (ours) & \tb{77.4} & \tb{87.7} & \tb{77.4} & \tb{78.5} \\
    & \citeauthor{tompson2015efficient} & 76.2 & 85.3 & 76.9 & 69.2 \\
    \midrule
    \multirow{2}{*}{AE} & \modelshort (ours) & \tb{1.16} & \tb{0.92} & \tb{1.14} & \tb{1.19} \\
    & \citeauthor{tompson2015efficient} & 1.20 & 1.02 & 1.19 & 1.54 \\
    \bottomrule
    \end{tabular}%}
\end{table}

\vpar{Results}
For 2D keypoint detection, we only train the keypoint estimator in our \modelshort (\fig{fig:architecture}a and \ref{fig:architecture}b) using the training images in each dataset. \fig{fig:human_keypoint_curves} shows the accuracy of keypoint estimation on the FLIC dataset. On this dataset, we employ a fine-level network for post-processing, as suggested by~\cite{tompson2015efficient}. Our method performs better than all previous methods~\citep{FLIC,tompson2014joint,tompson2015efficient,yang2011articulated,toshev2014deeppose} at all precisions. Moreover, the keypoint refinement step (\fig{fig:architecture}c) improves results significantly (about 2\% for a normalized distance $\geq0.15$), without which our framework has similar performance with \cite{tompson2015efficient}. Such improvement is also demonstrated in \fig{fig:sl}, where the heatmaps after refinement are far less noisy.

The accuracy of keypoint estimation on CUB-200-201 dataset is listed in \tbl{table:bird_keypoint_result}. Our method is better than~\cite{liu2013bird} in both metrics, and is comparable to the state-of-the-art~\citep{shih2015part}. Specifically, compared with~\cite{shih2015part}, our model more precisely estimates the keypoint locations for correctly detected parts (a lower AE), but misses more parts in the detection (a lower PCP). On our \datasetName dataset, our model achieves higher PCPs and lower AEs compared to the state-of-the-art~\citep{tompson2015efficient} for all categories, as shown in \tbl{table:other_keypoint}. These experiments in general demonstrate the effectiveness of our model on keypoint detection.

\begin{figure*}[t]
    \centering
    \begin{subfigure}{0.45\linewidth}
    \includegraphics[trim={0 1cm 15cm 0},clip,width=\linewidth]{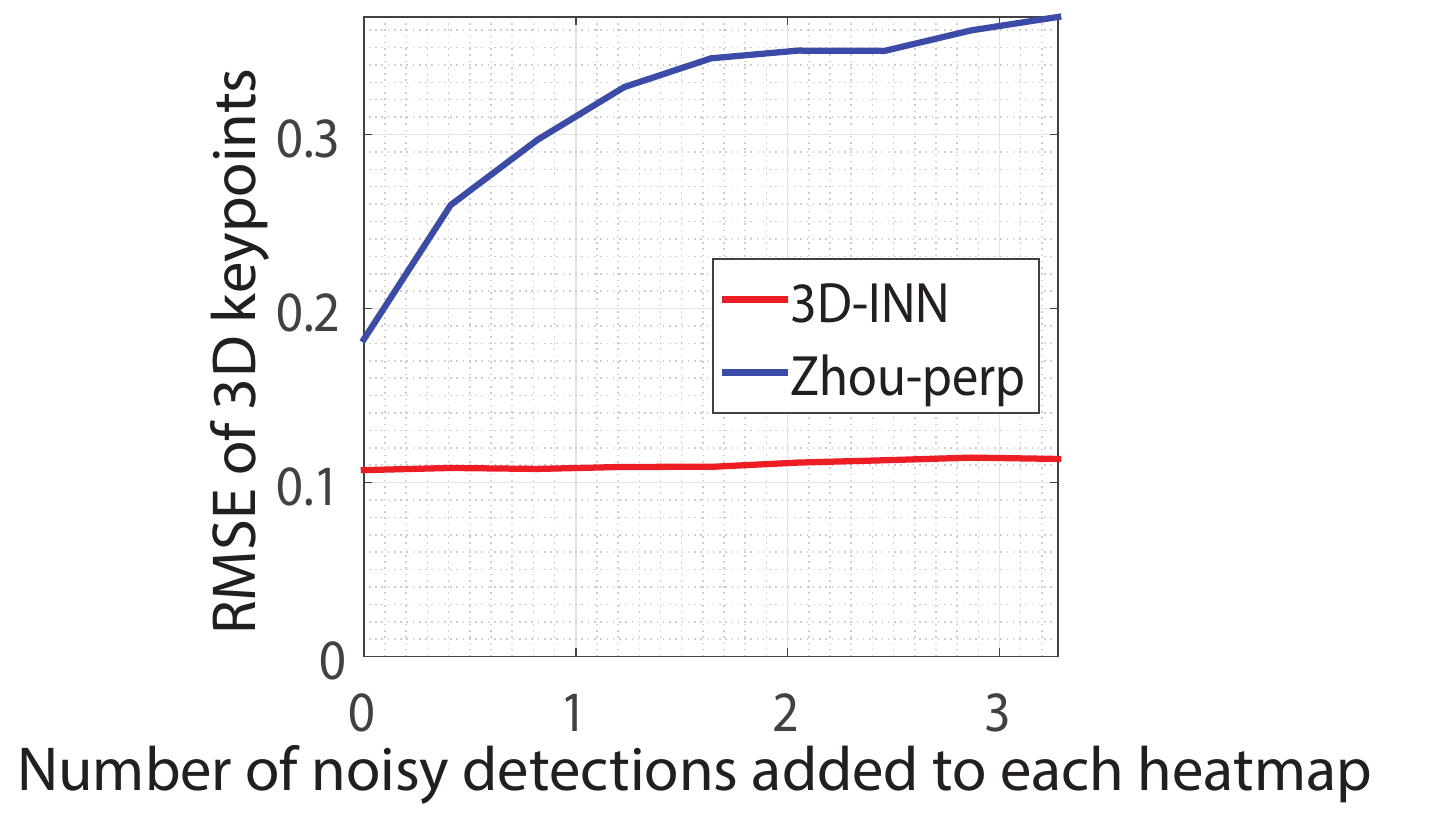}
    \caption{Structure estimation}
    \end{subfigure}
    \begin{subfigure}{0.45\linewidth}
    \includegraphics[trim={15cm 1cm 0 0},clip,width=\linewidth]{fig/ana.pdf}
    \caption{Viewpoint estimation}
    \end{subfigure}
    \caption{Plots comparing our method against an analytic solution on synthetic heatmap. 
    (a) The accuracy of 3D structure estimation; (b) The accuracy of 3D viewpoint estimation.
    }
    \label{fig:plot_syn}
\end{figure*}

\begin{figure*}[t]
\centering
\begin{subfigure}{0.35\linewidth}
\centering
\includegraphics[width=\linewidth]{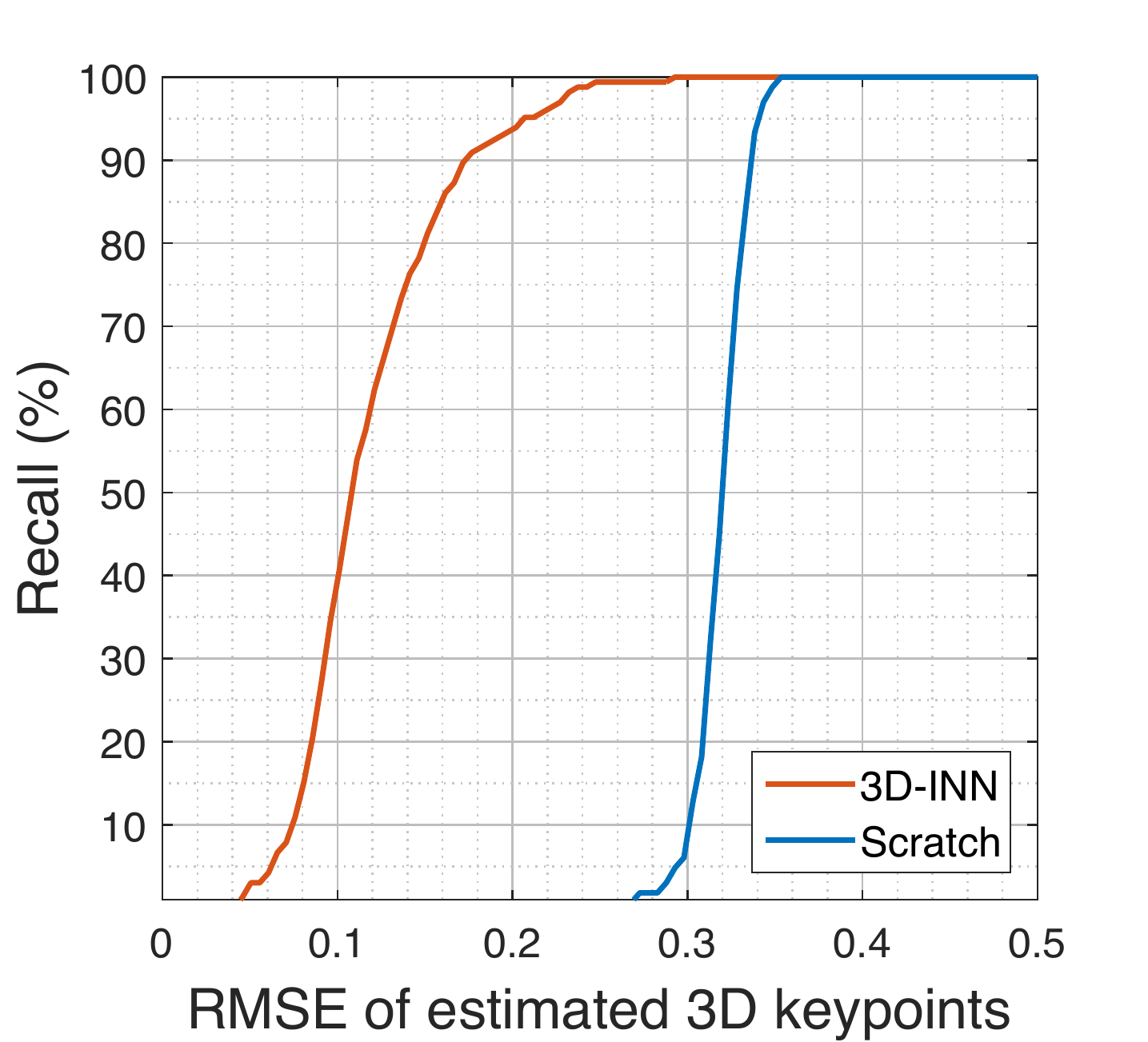}
\caption{Structure estimation}
\end{subfigure}
\begin{subfigure}{0.35\linewidth}
\centering
\includegraphics[width=\linewidth]{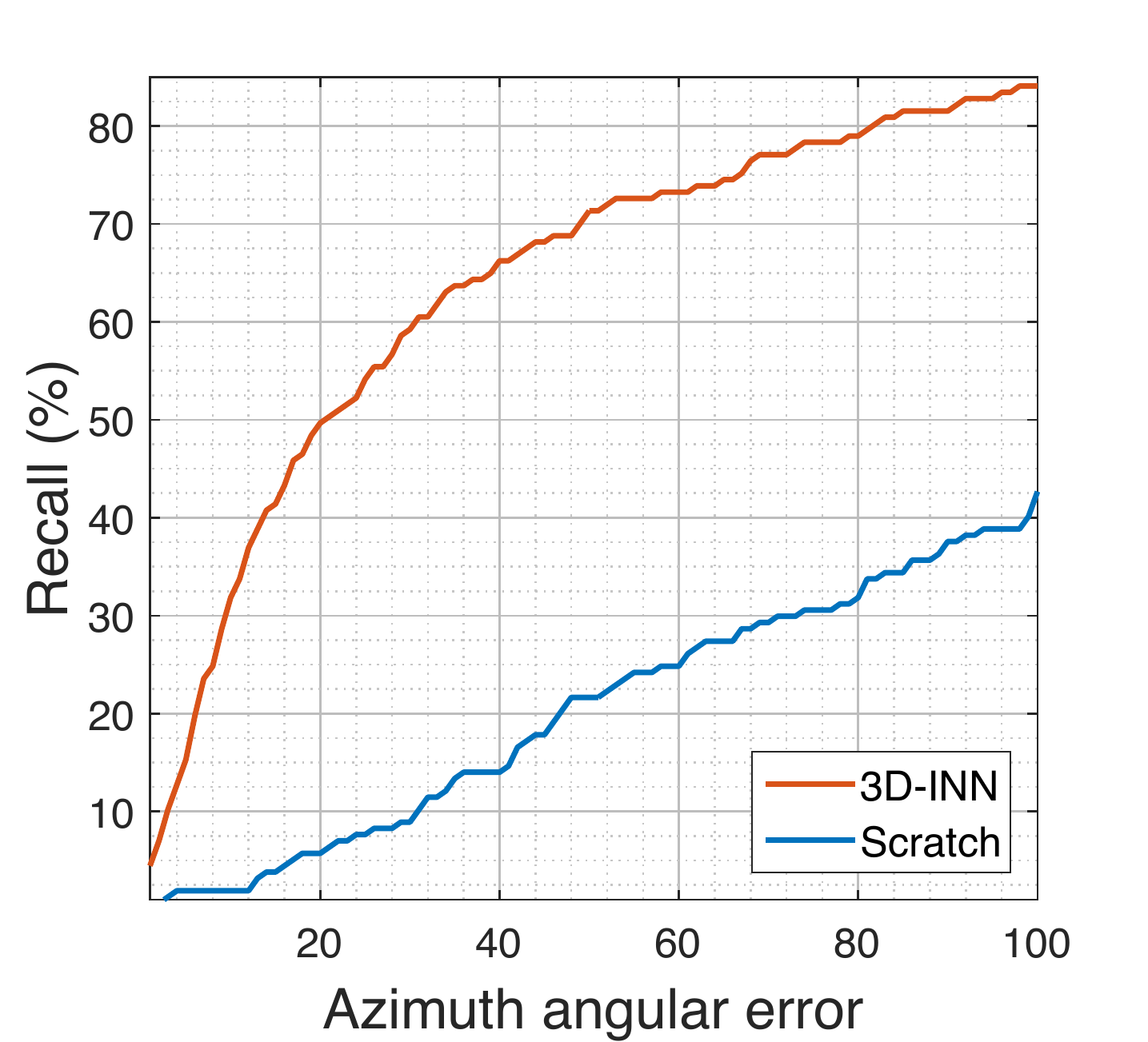}
\caption{Viewpoint estimation}
\end{subfigure}

\caption{Evaluation on chairs in the IKEA dataset~\citep{ikea}. The network trained with our paradigm (3D-INN) is significantly better than the one trained from \textit{scratch} on both 3D structure (a) and viewpoint estimation (b).    }
    \label{fig:train}
\end{figure*}

\subsection{3D Structure and Viewpoint Estimation}
\label{sec:eval_3d}

For 3D structural parameter estimation, we evaluate \modelshort from three different perspectives. First, we evaluate our 3D interpreter (\fig{fig:architecture}c alone) against the optimization-based method~\citep{zhou153d}. Second, we test our full pipeline on the IKEA dataset~\citep{ikea}, where ground truth 3D labels are available, comparing with both baselines and the state-of-the-art. Third, we show qualitative results on four datasets: \datasetName, IKEA, the SUN database~\citep{SUN}, and PASCAL 3D+~\cite{pascal3d}.

\vpar{Comparing with an optimization-based method.}
First, we compare our 3D interpreter (\fig{fig:architecture}c) with the state-of-the-art optimization-based method that directly minimizes re-projection error (\eqn{eq:pnp}) on the synthetic data. As most optimization based methods only consider the parallel projection, while we model perspective projection for real images, we extend the one by \cite{zhou153d} as follows: we first uses their algorithm to get an initial guess of internal parameters and viewpoints, and then applying a simple gradient descent method to refine it considering perspective distortion.

We generate synthetic data for this experiment, using the scheme described in \sect{sec:synthetic_data}. Each data point contains the 2D keypoint heatmaps of an object, and its corresponding 3D keypoint locations and viewpoint, which we would like to estimate. We also add different levels of salt-and-pepper noise to heatmaps to evaluate the robustness of both methods. We generated 30,000 training and 1,000 testing cases. Because the analytic solution only takes keypoint coordinates as input, we convert heatmaps to coordinates using an argmax function.

For both methods, we evaluate their performance on both 3D structure recovery and 3D viewpoint estimation. To evaluate the estimated 3D structure, we compare their accuracies on 3D keypoint estimation ($\cY$ in \sect{sec:skeleton}); for 3D viewpoint estimation, we compute errors in azimuth angle, following previous work~\citep{su15}. As the original algorithm by \cite{zhou153d} was mainly designed for the parallel projection and comparatively clean heatmaps, our 3D interpreter outperforms it in the presence of noise and perspective distortion, as shown in \fig{fig:plot_syn}. Our algorithm is also efficient, taking less than 50 milliseconds for each test image.

\begin{figure*}[t]
	\centering
    \includegraphics[width=\textwidth]{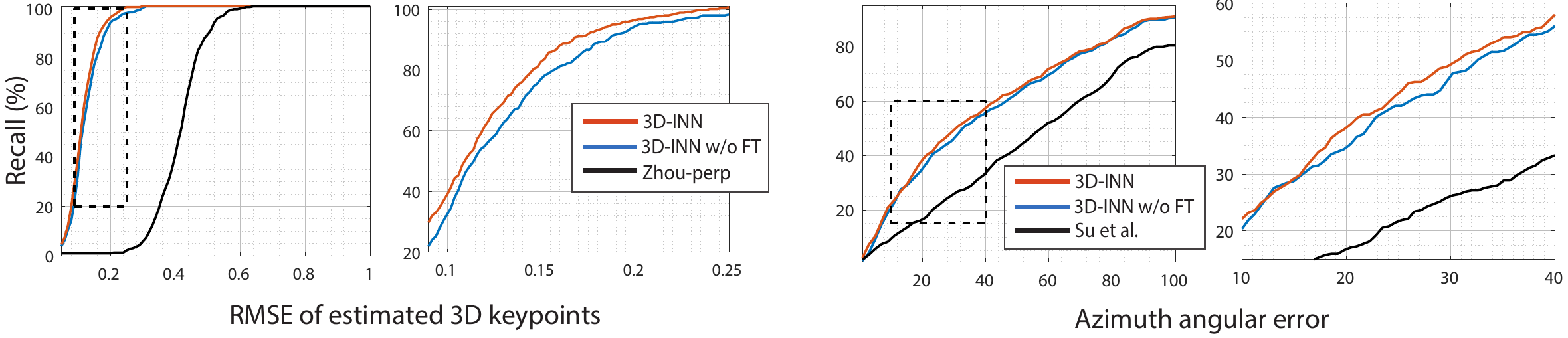} 
    \ \\
    \ \\
    
    \begin{subtable}{.48\linewidth}
    \centering 
    Average recall (\%) \\
    \begin{tabular}{lcccc} \toprule
Method& Bed& Sofa& Chair& Avg.\\ \midrule 
3D-INN & \textbf{88.64} & \textbf{88.03} & \textbf{87.84} & \textbf{88.03} \\ 
3D-INN w/o FT & 87.19 & 87.10 & 87.08 & 87.10 \\ 
\cite{zhou153d}-perp & 52.31 & 58.02 & 60.76 & 58.46 \\ 
\bottomrule \end{tabular}
    \caption{Structure estimation}
    \end{subtable} 
    \hfill
    \begin{subtable}{.48\linewidth}
    \centering 
    Average recall (\%) \\
    \begin{tabular}{lcccc} \toprule
Method& Table& Sofa& Chair& Avg.\\ \midrule 
3D-INN & \textbf{55.02} & 64.65 & \textbf{63.46} & \textbf{60.30} \\ 
3D-INN w/o FT & 52.33 & \textbf{65.45} & 62.01 & 58.90 \\ 
\cite{su15} & 52.73 & 35.65 & 37.69 & 43.34 \\ 
\bottomrule \end{tabular}
    \caption{Viewpoint estimation}
    \end{subtable} 
    
	\caption{\label{fig:ikea} Evaluation on the IKEA dataset~\citep{ikea}. (a) The accuracy of structure estimation. RMSE-Recall curved is shown in the first row, and zoomed-views of the dashed rectangular regions are shown on the right. The third row shows the average recall on all thresholds. (b) The accuracy of viewpoint estimation. %Zhou-perp is a modified algorithm based on~\cite{zhou153d} to deal with perspective distortion.
	}
	%\vspace{-5pt}
\end{figure*}
\begin{table*}[t!]
\footnotesize
\centering
\caption{
\label{table:pascal3d}Joint object detection and viewpoint estimation on PASCAL 3D+~\citep{pascal3d}. Following previous work, we use Average Viewpoint Precision (AVP) as our measure, which extends AP so that a true positive should have both a correct bounding box and a correct viewpoint (here we use a 4-view quantization). Both \modelshort and V\&K~\citep{tulsiani2015viewpoints} use R-CNN~\citep{RCNN} for object detection, precluding the influence of object detectors. The others use their own detection algorithm. VDPM~\citep{pascal3d} and DPM-VOC+VP~\citep{pepik2012teaching} are trained on PASCAL VOC 2012, V\&K~\citep{tulsiani2015viewpoints} is trained on PASCAL 3D+, \cite{su15} is trained on PASCAL VOC 2012, together with synthetic 3D CAD models, and \modelshort is trained on \datasetName. }
\begin{tabular}{lccccc}
\toprule
Category & VDPM~\citep{pascal3d} & DPM-VOC+VP~\citep{pepik2012teaching} & \cite{su15} & V\&K~\citep{tulsiani2015viewpoints} & 3D-INN \\ 
\midrule
Chair & 6.8 & 6.1 & 15.7 & \textbf{25.1} & 23.1 \\
Sofa & 5.1 & 11.8 & 18.6 & 43.8 & \textbf{45.8} \\
Car & 20.2 & 36.9 & 41.8 & \textbf{55.2} & 52.2 \\
\bottomrule
\end{tabular}
\end{table*}

\begin{figure*}[t!]
	\centering
	\includegraphics[width=.95\linewidth]{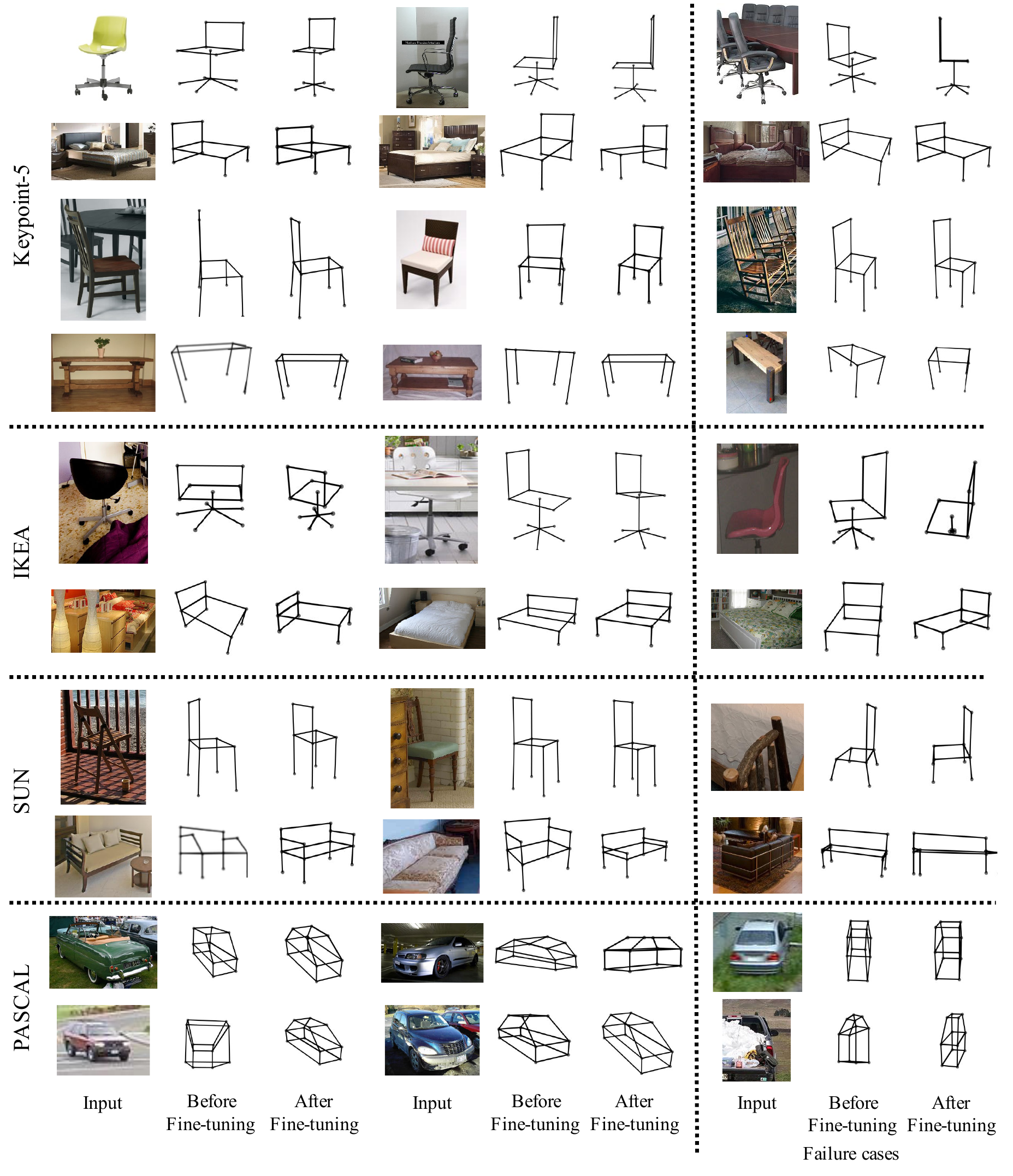}
	\caption{\label{fig:res} Qualitative results on \datasetName, IKEA, and SUN databases. For each example, the first one is the input image, the second one is the reconstruct 3D skeleton using the network before fine-tuning, and third one is using the network after fine-tuning. The last column shows failure cases.}
\end{figure*}

\begin{figure*}
    \centering
    \includegraphics[width=\linewidth]{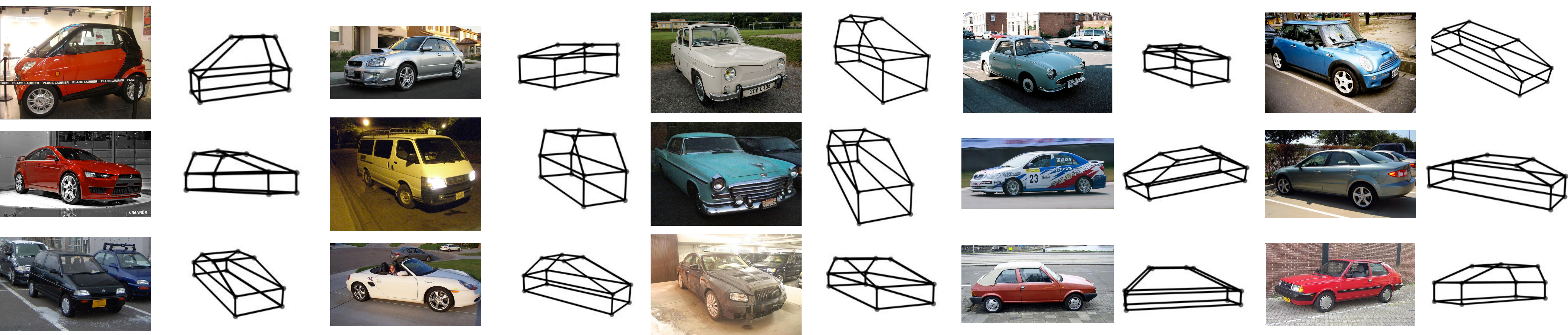}
    \caption{Car structure estimation on images from the PASCAL 3D+ dataset}
    \label{fig:car}
\end{figure*}
\begin{figure*}[t]
    \centering

    \begin{subfigure}{0.49\linewidth}
    \centering
    \includegraphics[width=\linewidth]{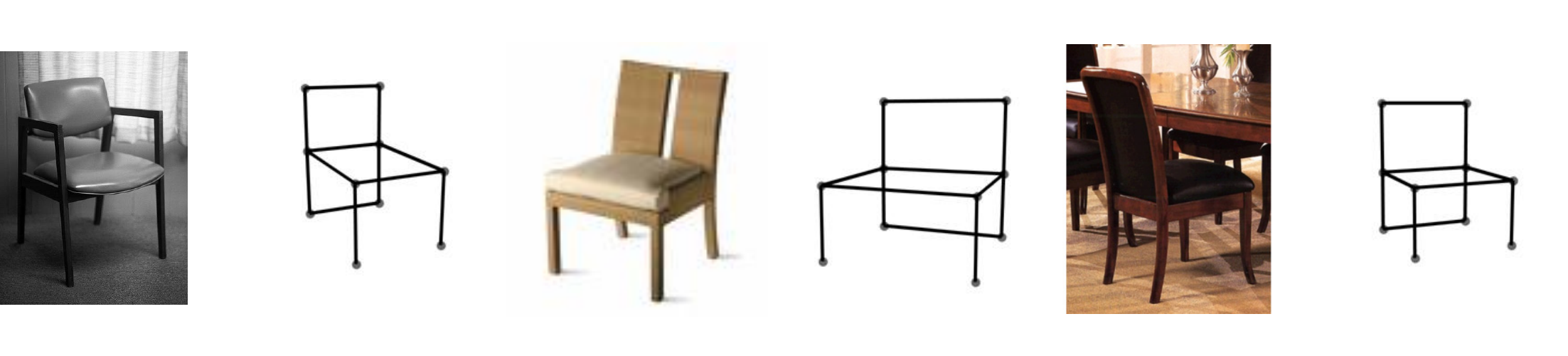} 
    \caption{Training: beds, Test: chairs}
    \end{subfigure}
    \hfill
    \begin{subfigure}{0.49\linewidth}
    \centering
    \includegraphics[width=\linewidth]{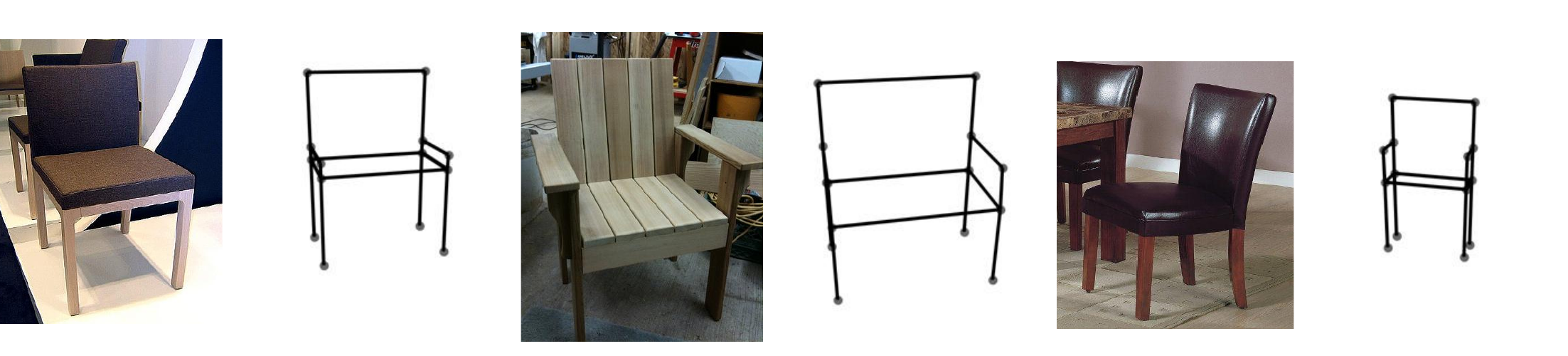}
    \caption{Training: sofas, Test: chairs}
    \end{subfigure}
    
    \caption{Qualitative results on chairs using networks trained on sofas or beds. In most cases models provide reasonable output. Mistakes are often due to the difference between the training and test sets, \eg, in the third example, the model trained on beds fails to estimate chairs facing backward.}
    \label{fig:cross_category}
\end{figure*}

\vpar{Evaluating the full pipeline.}
We now evaluate \modelshort on estimating 3D structure and 3D viewpoint. We use the IKEA dataset~\citep{ikea} for evaluation, as it provides ground truth 3D mesh models and the associated viewpoints for testing images. We manually label ground truth 3D keypoint locations on provided 3D meshes, and calculate the root-mean-square error (RMSE) between estimated and ground truth 3D keypoint locations. 

As IKEA only has no more than 200 images per category, we instead train \modelshort on our \datasetName, as well as one million synthetic data points, using the strategy described in \sect{sec:synthetic_data}. Note that, first, we are only using no more than 2,000 real images per category for training and, second, we are testing the trained model on different datasets, avoiding possible dataset biases~\citep{datasetbias}.

We first compare with a baseline method to evaluate our training paradigm: we show quantitative comparisons between \modelshort trained using our paradigm proposed in \sect{sec:synthetic_data}, and the same network but only end-to-end trained with real images, without having the two pre-training stages. We called it the \textit{scratch} model.

As shown in the RMSE-Recall curve in \fig{fig:train}, \modelshort performs much better than \textit{scratch} on both 3D structure and viewpoint estimation. The average recall of \modelshort is about 20\% higher than \textit{scratch} in 3D structure estimation, and about 40\% higher in 3D pose estimation. This shows the effectiveness of the proposed training paradigm.

We then compare our full model with the state-of-the-art methods. The left half of \fig{fig:ikea} shows RMSE-Recall curve of both our algorithm and the optimization-based method described above (\cite{zhou153d}-perp). The $y$-axis shows the recall --- the percentage of testing samples under a certain RMSE threshold. We test two versions of our algorithm: with fine-tuning (3D-INN) and without fine-tuning (3D-INN w/o FT). Both significantly outperform the optimization-based method~\citep{zhou153d}. This is because the method from \cite{zhou153d} was not designed to handle noisy keypoint estimation and perspective distortions, while our \modelshort can deal with them. Also, fine-tuning improves the accuracy of keypoint estimation by about 5\% under the RMSE threshold 0.15.

Though we focus on recovering 3D object structure, as an extension, we also evaluate \modelshort on 3D viewpoint estimation. We compare it with the state-of-the-art viewpoint estimation algorithm by \cite{su15}. The right half of \fig{fig:ikea} shows the results (recall) in azimuth angle. As shown in the table, \modelshort outperforms \cite{su15} by about $40\%$ (relative), measured in average recall. This is mainly because it is not straightforward for \cite{su15}, mostly trained on (cropped) synthesized images, to deal with the large number of heavily occluded objects in the IKEA dataset.

Although our algorithm assumes a centered object in an input image, we can apply it, in combination with an object detection algorithm, on images where object locations are unknown. We evaluate the results of joint object detection and viewpoint estimation on PASCAL 3D+ dataset~\citep{pascal3d}. PASCAL 3D+ and Keypoint-5 has two overlapping categories: chair and sofa, and we evaluate on both. We also study an additional object category, car, for which \modelshort is trained on 1,000 car images from ImageNet~\citep{russakovsky2015imagenet} with 2D keypoint annotations. Following \cite{tulsiani2015viewpoints}, we use non-occluded and non-truncated objects for testing. We use the standard R-CNN~\citep{RCNN} for object detection, and our \modelshort for viewpoint estimation. 

\tbl{table:pascal3d} shows that \modelshort is comparable with Viewpoints and Keypoints (V\&K by \cite{tulsiani2015viewpoints}), and outperforms other algorithms with a significant margin. Both \modelshort and V\&K use R-CNN~\citep{RCNN} for object detection (we use the R-CNN detection results provided by \cite{tulsiani2015viewpoints}); this rules out the influence of object detectors. Further, while all the other algorithms are trained on either PASCAL VOC or PASCAL 3D+, ours is trained on \datasetName or ImageNet. This indicates our learned model transfers well across datasets, without suffering much from the domain adaptation issue. 

\begin{figure*}[t]
    \centering
    \includegraphics[width=0.8\linewidth]{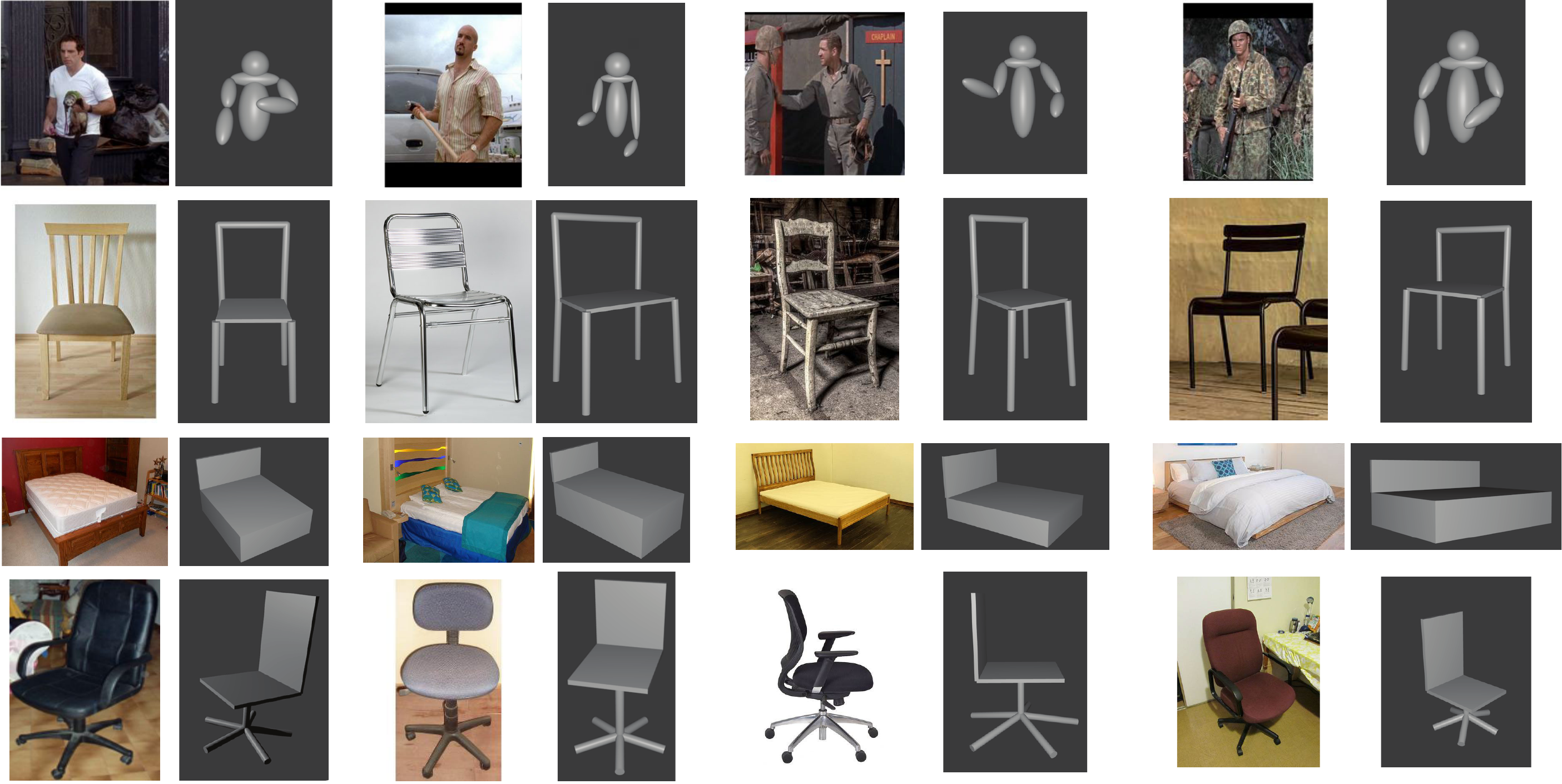}
    \caption{Visualization of 3D reconstruction results. We render objects using \textit{Blender}.}
    \label{fig:render}
\end{figure*}

\begin{figure*}[t]
    \centering
    
    \begin{subfigure}{0.53\linewidth}
    \includegraphics[trim={0 1.2cm 20cm 0},clip,width=\linewidth]{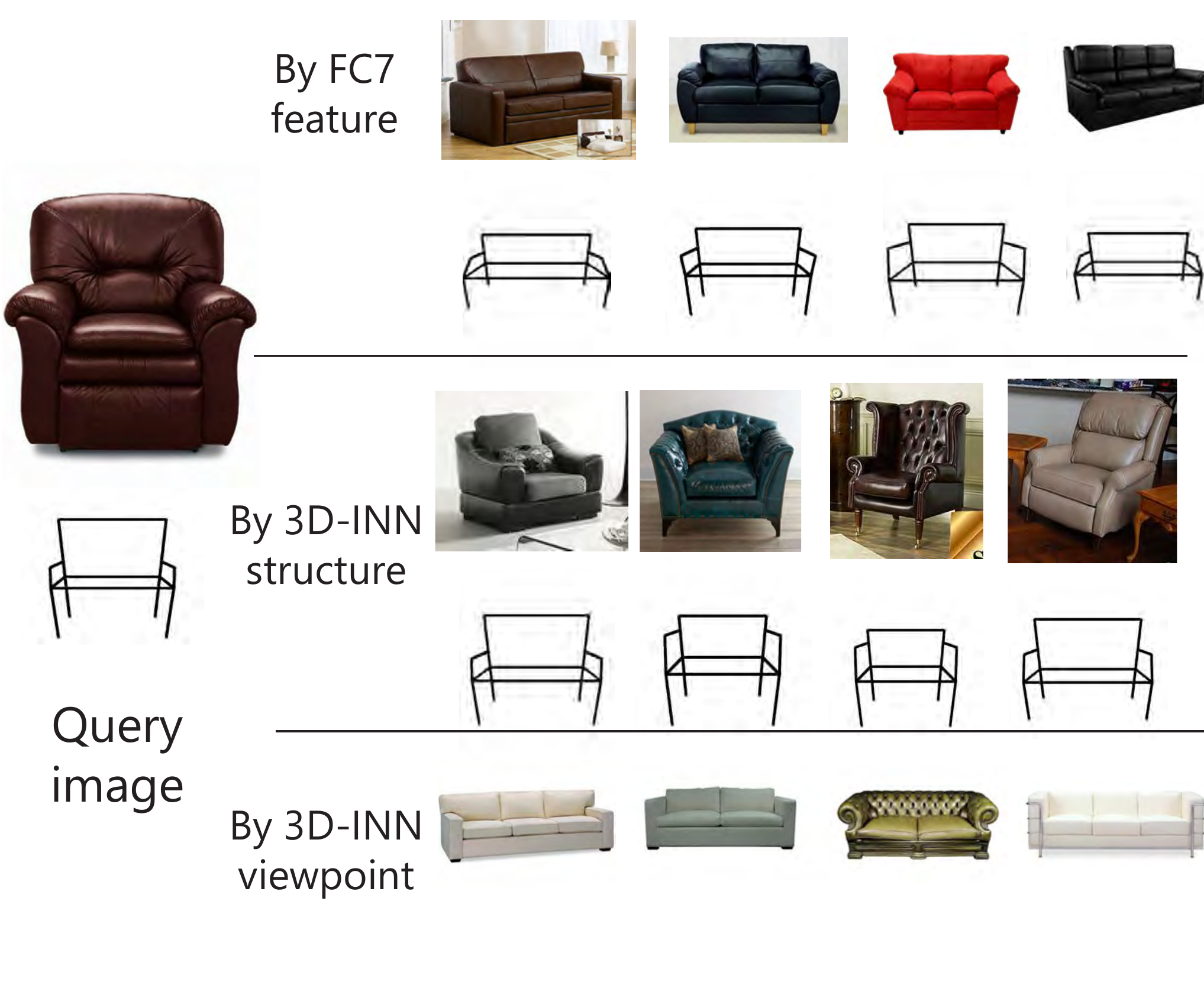}
    \caption{Retrieval results of a sofa image}
    \end{subfigure}
    \hfill
    \begin{subfigure}{0.46\linewidth}
    \includegraphics[trim={22cm 1.2cm 0 0},clip,width=\linewidth]{fig/retrieval.pdf}
    \caption{Retrieval results of a chair image}
    \end{subfigure}
    
    \caption{Retrieval results for a sofa (b) and a chair (b) in different feature spaces. \modelshort helps to retrieve objects with similar 3D structure or pictured in a similar viewpoint.} 
    \label{fig:retrieval}
\end{figure*}

\begin{figure*}
    \centering
    \includegraphics[width=0.6\linewidth]{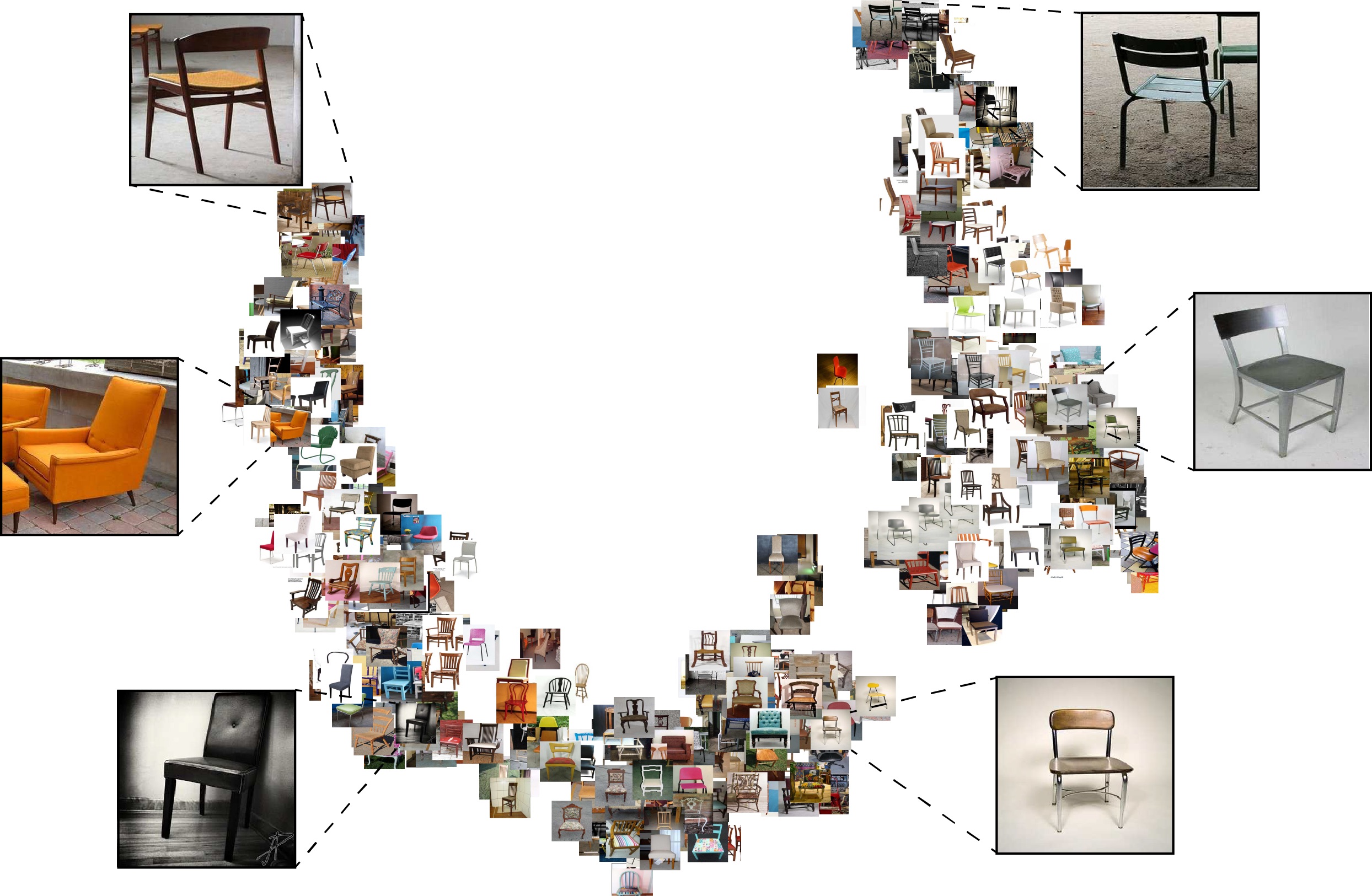}
    \caption{Object graph visualization based on learned object representations: we visualize images using t-SNE~\citep{tsne} on 3D viewpoints predicted by \modelshort.}
    \label{fig:objgraph}
\end{figure*}

\vpar{Qualitative results on benchmarks.}
We now show qualitative results on \datasetName, IKEA, the SUN database \citep{SUN}, and the PASCAL 3D+ dataset \citep{pascal3d} in \fig{fig:res}. When the image is clean and objects are not occluded, our algorithm can recover 3D object structure and viewpoint with high accuracy. Fine-tuning further helps to improve the results (see chairs at row 1 column 1, and row 4 column 1). Our algorithm is also robust to partial occlusion, demonstrated by the IKEA bed at row 5 column 1. We show failure cases in the last column: one major failure case is when the object is heavily cropped in the input image (the last column, row 4 to 7), as the 3D object skeleton becomes hard to infer. \fig{fig:car} shows more results on car structure recovery.

When \modelshort is used in combination with detection models, it needs to deal with imperfect detection results. Here, we also evaluate \modelshort on noisy input, specifically, on images with an object from a different but similar category. \fig{fig:cross_category} shows the recovered 3D structures of chairs using a model trained either on sofas or beds. In most cases \modelshort still provides reasonable output, and the mistakes are mostly due to the difference between training and test sets, \eg, the model trained on beds does not perform well on chairs facing backward, because there are almost no beds with a similar viewpoint in the training set.

\begin{figure*}[t!]
\centering
\includegraphics[width=.94\linewidth]{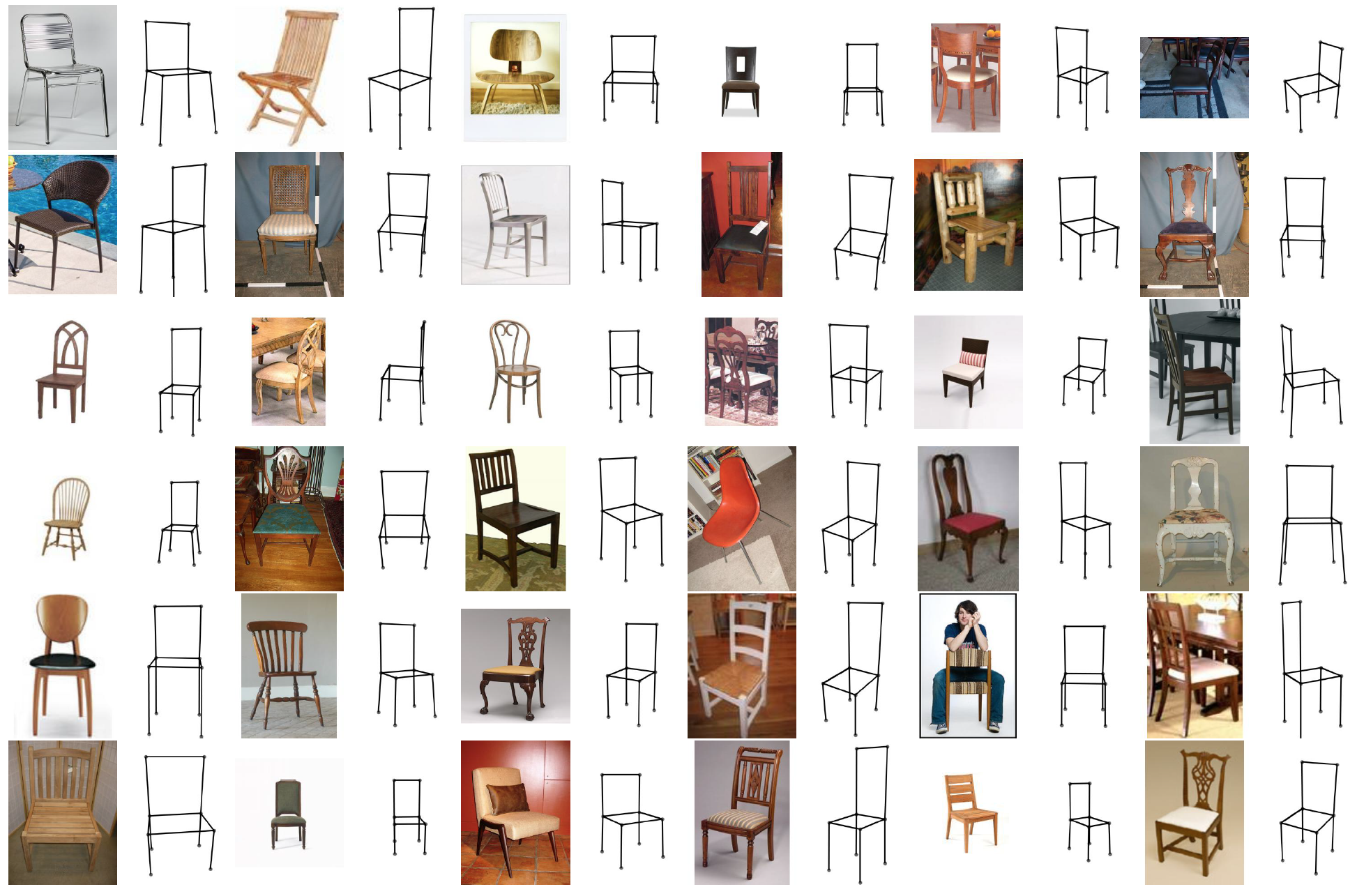}
\caption{Estimated 3D skeletons on more \datasetName chair images. Images are randomly sampled from the test set.}
\label{fig:k5chair}
\end{figure*}

\begin{figure*}[t!]
\centering
\includegraphics[width=.94\linewidth]{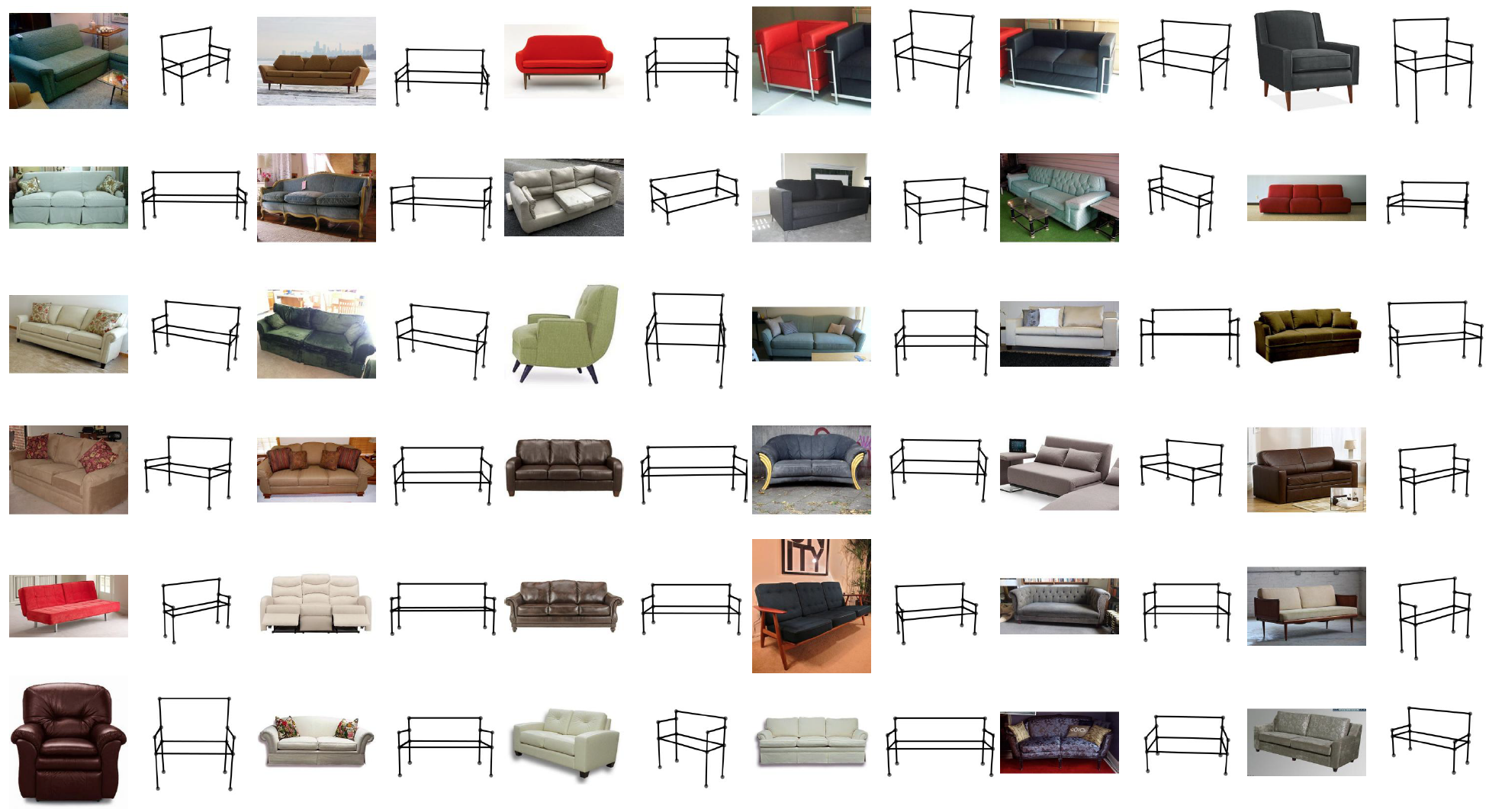}
\caption{Estimated 3D skeletons on more \datasetName sofa images. Images are randomly sampled from the test set.}
\label{fig:k5sofa}
\end{figure*}

At the end of the manuscript, we supply more results on chair and sofa images randomly sampled from the test set of \datasetName. \fig{fig:k5chair} and \fig{fig:k5sofa} show the estimated skeletons for chairs and sofas, respectively.
\section{Applications}

The inferred latent parameters, as a compact and informative representation of objects in images, have wide applications. In this section, we demonstrate representative ones including 3D object rendering, image retrieval, and object graph visualization.

\vpar{3D Object Rendering}
Given an estimated 3D object structure, we can render it in a 3D graphics engine like Blender, as shown in \fig{fig:render}.

\vpar{Image Retrieval}
Using estimated 3D structural and viewpoint information, we can retrieve images based on their 3D configurations. \fig{fig:retrieval} shows image retrieval results using FC7 features from AlexNet~\citep{alexnet} and using the 3D structure and viewpoint learned by \modelshort. Our retrieval database includes all testing images of chairs and sofas in \datasetName. In each row, we sort the best matches of the query image, measured by Euclidean distance in a specific feature space. We retrieve images in two ways: \emph{by structure} uses estimated internal structural parameters ($\{\alpha_i\}$ in \eqn{eq:pnp}), and \emph{by viewpoint} uses estimated external viewpoint parameters ($R$ in \eqn{eq:pnp}).

\vpar{Object Graph}
Similar to the retrieval task, we visualize all test images for chairs in \datasetName in \fig{fig:objgraph}, using t-SNE~\citep{tsne} on estimated 3D viewpoints. Note the smooth transition from the chairs facing left to those facing right. 
\section{Discussions}
\label{sec:discuss}

We have introduced \model (\modelshort). From a single image, our model recovers the 2D keypoints and 3D structure of a (possibly deformable) object, as well as camera parameters. To achieve this goal, we used 3D skeletons as an abstract 3D representation, incorporated a projection layer to the network for learning 3D parameters from 2D labels, and employed keypoint heatmaps to connect real and synthetic data. Empirically, we showed that \modelshort performs well on both 2D keypoint estimation and 3D structure and viewpoint recovery, comparable to or better than the state of the art. Further, various applications demonstrated the potential of the skeleton representation learned by \modelshort. 

We choose to model objects via 3D skeletons and the corresponding 2D keypoints, as opposed to other dense 3D representations such as voxels, meshes, and point clouds, because skeletons offer unique advantages. First, given an RGB image, its sparse 2D annotations like keypoints are easier and more affordable to acquire, and can be used as 2D supervision for 3D skeleton and viewpoint estimation; in comparison, it is prohibitively challenging to obtain dense annotations like a depth map to constrain 3D reconstructions in voxels or meshes. Second, the employed base shapes carry rich category-specific shape priors, with which \modelshort can encode an object skeleton with a few parameters. This feature is particularly useful on platforms with severe memory and computational constraints, such as on autonomous cars and on mobile phones.

That being said, skeletons have their own limitations. The most significant is on its generalization power: there are many real-world objects whose keypoints are hard to define, such as trees, flowers, and deformable shapes like ropes; in those cases, there lacks a straightforward way to apply \modelshort to model these objects. Recent research on 3D reconstruction via richer, generic intermediate representations like intrinsic images~\citep{barrow} suggests a potential solution to the problem, though as discussed above it is much harder to obtain annotated intrinsic images, compared to keypoints~\citep{marrnet}.

In this work, we focus on single-view 3D reconstruction. As discussed in \sect{sec:intro}, requiring only a single image as input has unique practical advantages, in addition to its scientific value. First, our algorithm can be directly applied to cases where only in-the-wild images are available, not multi-view images or videos. Second, taking a single image as input enables online inference and therefore fits real-time applications; in contrast, most multi-view reconstruction algorithms are offline. It is also possible to our \modelshort to use multi-view data when they are available~\citep{kar2017learning}, and more generally, to integrate viewer-centered and object-centered representations in a principled manner~\citep{hinton1981parallel}.

\modelshort estimates the 3D skeleton and pose of an object from an RGB image, and can therefore be applied to the enormous existing RGB data. But we are also aware that depth sensors have recently become affordable to end users~\citep{newcombe2011kinectfusion}, and large-scale RGB-D datasets are being built~\citep{song2016semantic,mccormac2017scenenet}. Depth data help to resolve the ambiguity in the projection from 3D shapes to 2D images, allow object structure prediction in metric scale, and enable wide applications~\citep{chen2012kinetre}.  Hence, a promising future research topic would be to extend the current framework to handle depth data, while enforcing the 2D-3D differentiable consistencies in various forms~\citep{tulsiani2017multi,marrnet}.

\section*{Acknowledgement} This work is supported by NSF Robust Intelligence 1212849 and NSF Big Data 1447476 to W.F., NSF Robust Intelligence 1524817 to A.T., ONR MURI N00014-16-1-2007 to J.B.T., Shell Research, the Toyota Research Institute, and the Center for Brain, Minds and Machines (NSF STC award CCF-1231216). The authors would like to thank Nvidia for GPU donations. Part of this work was done when Jiajun Wu was an intern at Facebook AI Research, and Tianfan Xue was a graduate student at MIT CSAIL.

\bibliographystyle{spbasic}      % basic style, author-year citations
\bibliography{3dinn}   % name your BibTeX data base

\begin{thebibliography}{73}
\providecommand{\natexlab}[1]{#1}
\providecommand{\url}[1]{{#1}}
\providecommand{\urlprefix}{URL }
\expandafter\ifx\csname urlstyle\endcsname\relax
  \providecommand{\doi}[1]{DOI~\discretionary{}{}{}#1}\else
  \providecommand{\doi}{DOI~\discretionary{}{}{}\begingroup
  \urlstyle{rm}\Url}\fi
\providecommand{\eprint}[2][]{\url{#2}}

\bibitem[{Akhter and Black(2015)}]{akhter2015pose}
Akhter I, Black MJ (2015) Pose-conditioned joint angle limits for 3d human pose
  reconstruction. In: IEEE Conference on Computer Vision and Pattern
  Recognition

\bibitem[{Aubry et~al(2014)Aubry, Maturana, Efros, Russell, and
  Sivic}]{Aubry14}
Aubry M, Maturana D, Efros A, Russell B, Sivic J (2014) Seeing 3d chairs:
  exemplar part-based 2d-3d alignment using a large dataset of cad models. In:
  IEEE Conference on Computer Vision and Pattern Recognition

\bibitem[{Bansal and Russell(2016)}]{bansal2016marr}
Bansal A, Russell B (2016) Marr revisited: 2d-3d alignment via surface normal
  prediction. In: IEEE Conference on Computer Vision and Pattern Recognition

\bibitem[{Barrow and Tenenbaum(1978)}]{barrow}
Barrow HG, Tenenbaum JM (1978) Recovering intrinsic scene characteristics from
  images. Computer vision systems

\bibitem[{Belhumeur et~al(2013)Belhumeur, Jacobs, Kriegman, and
  Kumar}]{consensus}
Belhumeur PN, Jacobs DW, Kriegman DJ, Kumar N (2013) Localizing parts of faces
  using a consensus of exemplars. IEEE Transactions on Pattern Analysis and
  Machine intelligence 35(12):2930--2940

\bibitem[{Bever and Poeppel(2010)}]{bever2010analysis}
Bever TG, Poeppel D (2010) Analysis by synthesis: a (re-) emerging program of
  research for language and vision. Biolinguistics 4(2-3):174--200

\bibitem[{Bourdev et~al(2010)Bourdev, Maji, Brox, and Malik}]{poselet}
Bourdev L, Maji S, Brox T, Malik J (2010) Detecting people using mutually
  consistent poselet activations. In: European Conference on Computer Vision

\bibitem[{Carreira et~al(2016)Carreira, Agrawal, Fragkiadaki, and
  Malik}]{carreira2015human}
Carreira J, Agrawal P, Fragkiadaki K, Malik J (2016) Human pose estimation with
  iterative error feedback. In: IEEE Conference on Computer Vision and Pattern
  Recognition

\bibitem[{Chen et~al(2012)Chen, Izadi, and Fitzgibbon}]{chen2012kinetre}
Chen J, Izadi S, Fitzgibbon A (2012) Kin{\^e}tre: animating the world with the
  human body. In: ACM symposium on User Interface Software and Technology

\bibitem[{Choy et~al(2016)Choy, Xu, Gwak, Chen, and Savarese}]{choy20163d}
Choy CB, Xu D, Gwak J, Chen K, Savarese S (2016) 3d-r2n2: A unified approach
  for single and multi-view 3d object reconstruction. In: European Conference
  on Computer Vision

\bibitem[{Dosovitskiy et~al(2015)Dosovitskiy, Tobias~Springenberg, and
  Brox}]{DB15}
Dosovitskiy A, Tobias~Springenberg J, Brox T (2015) Learning to generate chairs
  with convolutional neural networks. In: IEEE Conference on Computer Vision
  and Pattern Recognition

\bibitem[{Fidler et~al(2012)Fidler, Dickinson, and Urtasun}]{urtasun_3d_car}
Fidler S, Dickinson SJ, Urtasun R (2012) 3d object detection and viewpoint
  estimation with a deformable 3d cuboid model. In: Advances in Neural
  Information Processing Systems

\bibitem[{Girshick et~al(2014)Girshick, Donahue, Darrell, and Malik}]{RCNN}
Girshick R, Donahue J, Darrell T, Malik J (2014) Rich feature hierarchies for
  accurate object detection and semantic segmentation. In: IEEE Conference on
  Computer Vision and Pattern Recognition

\bibitem[{Hejrati and Ramanan(2012)}]{hejrati2012analyzing}
Hejrati M, Ramanan D (2012) Analyzing 3d objects in cluttered images. In:
  Advances in Neural Information Processing Systems

\bibitem[{Hejrati and Ramanan(2014)}]{synthesis3d}
Hejrati M, Ramanan D (2014) Analysis by synthesis: 3d object recognition by
  object reconstruction. In: IEEE Conference on Computer Vision and Pattern
  Recognition

\bibitem[{Hinton and Ghahramani(1997)}]{hinton1997generative}
Hinton GE, Ghahramani Z (1997) Generative models for discovering sparse
  distributed representations. Philosophical Transactions of the Royal Society
  of London B: Biological Sciences 352(1358):1177--1190

\bibitem[{Hinton(1981)}]{hinton1981parallel}
Hinton GF (1981) A parallel computation that assigns canonical object-based
  frames of reference. In: International Joint Conference on Artificial
  Intelligence

\bibitem[{Hu and Zhu(2015)}]{hu2015learning}
Hu W, Zhu SC (2015) Learning 3d object templates by quantizing geometry and
  appearance spaces. IEEE Transactions on Pattern Analysis and Machine
  intelligence 37(6):1190--1205

\bibitem[{Huang et~al(2015)Huang, Wang, and Koltun}]{huang2015single}
Huang Q, Wang H, Koltun V (2015) Single-view reconstruction via joint analysis
  of image and shape collections. ACM Transactions on Graphics 34(4):87

\bibitem[{Jaderberg et~al(2015)Jaderberg, Simonyan, Zisserman, and
  Kavukcuoglu}]{jaderberg2015spatial}
Jaderberg M, Simonyan K, Zisserman A, Kavukcuoglu K (2015) Spatial transformer
  networks. In: Advances in Neural Information Processing Systems

\bibitem[{Kar et~al(2015)Kar, Tulsiani, Carreira, and Malik}]{kar2015category}
Kar A, Tulsiani S, Carreira J, Malik J (2015) Category-specific object
  reconstruction from a single image. In: IEEE Conference on Computer Vision
  and Pattern Recognition

\bibitem[{Kar et~al(2017)Kar, H{\"a}ne, and Malik}]{kar2017learning}
Kar A, H{\"a}ne C, Malik J (2017) Learning a multi-view stereo machine. In:
  Advances in Neural Information Processing Systems

\bibitem[{Krizhevsky et~al(2012)Krizhevsky, Sutskever, and Hinton}]{alexnet}
Krizhevsky A, Sutskever I, Hinton GE (2012) Imagenet classification with deep
  convolutional neural networks. In: Advances in Neural Information Processing
  Systems

\bibitem[{Kulkarni et~al(2015{\natexlab{a}})Kulkarni, Kohli, Tenenbaum, and
  Mansinghka}]{kulkarni2015picture}
Kulkarni TD, Kohli P, Tenenbaum JB, Mansinghka V (2015{\natexlab{a}}) Picture:
  A probabilistic programming language for scene perception. In: IEEE
  Conference on Computer Vision and Pattern Recognition

\bibitem[{Kulkarni et~al(2015{\natexlab{b}})Kulkarni, Whitney, Kohli, and
  Tenenbaum}]{kulkarni2015deep}
Kulkarni TD, Whitney WF, Kohli P, Tenenbaum JB (2015{\natexlab{b}}) Deep
  convolutional inverse graphics network. In: Advances in Neural Information
  Processing Systems

\bibitem[{Leclerc and Fischler(1992)}]{leclerc1992optimization}
Leclerc YG, Fischler MA (1992) An optimization-based approach to the
  interpretation of single line drawings as 3d wire frames. International
  Journal of Computer Vision 9(2):113--136

\bibitem[{Li et~al(2015)Li, Su, Qi, Fish, Cohen-Or, and Guibas}]{li2015joint}
Li Y, Su H, Qi CR, Fish N, Cohen-Or D, Guibas LJ (2015) Joint embeddings of
  shapes and images via cnn image purification. ACM Transactions on Graphics
  34(6):234

\bibitem[{Lim et~al(2013)Lim, Pirsiavash, and Torralba}]{ikea}
Lim JJ, Pirsiavash H, Torralba A (2013) Parsing ikea objects: Fine pose
  estimation. In: IEEE International Conference on Computer Vision

\bibitem[{Lim et~al(2014)Lim, Khosla, and Torralba}]{fpm}
Lim JJ, Khosla A, Torralba A (2014) {FPM}: Fine pose parts-based model with 3d
  cad models. In: European Conference on Computer Vision

\bibitem[{Liu and Belhumeur(2013)}]{liu2013bird}
Liu J, Belhumeur PN (2013) Bird part localization using exemplar-based models
  with enforced pose and subcategory consistency. In: IEEE International
  Conference on Computer Vision

\bibitem[{Lowe(1987)}]{lowe1987three}
Lowe DG (1987) Three-dimensional object recognition from single two-dimensional
  images. Artificial intelligence 31(3):355--395

\bibitem[{Van~der Maaten and Hinton(2008)}]{tsne}
Van~der Maaten L, Hinton G (2008) Visualizing data using t-sne. Journal of
  Machine Learning Research 9(11):2579--2605

\bibitem[{McCormac et~al(2017)McCormac, Handa, Leutenegger, and
  Davison}]{mccormac2017scenenet}
McCormac J, Handa A, Leutenegger S, Davison AJ (2017) Scenenet rgb-d: Can 5m
  synthetic images beat generic imagenet pre-training on indoor segmentation.
  In: IEEE International Conference on Computer Vision

\bibitem[{Newcombe et~al(2011)Newcombe, Izadi, Hilliges, Molyneaux, Kim,
  Davison, Kohi, Shotton, Hodges, and Fitzgibbon}]{newcombe2011kinectfusion}
Newcombe RA, Izadi S, Hilliges O, Molyneaux D, Kim D, Davison AJ, Kohi P,
  Shotton J, Hodges S, Fitzgibbon A (2011) Kinectfusion: Real-time dense
  surface mapping and tracking. In: IEEE International Symposium on Mixed and
  Augmented Reality, pp 127--136

\bibitem[{Newell et~al(2016)Newell, Yang, and Deng}]{newell2016stacked}
Newell A, Yang K, Deng J (2016) Stacked hourglass networks for human pose
  estimation. In: European Conference on Computer Vision

\bibitem[{Pepik et~al(2012)Pepik, Stark, Gehler, and
  Schiele}]{pepik2012teaching}
Pepik B, Stark M, Gehler P, Schiele B (2012) Teaching 3d geometry to deformable
  part models. In: IEEE Conference on Computer Vision and Pattern Recognition

\bibitem[{Prasad et~al(2010)Prasad, Fitzgibbon, Zisserman, and
  Van~Gool}]{prasad2010finding}
Prasad M, Fitzgibbon A, Zisserman A, Van~Gool L (2010) Finding nemo: Deformable
  object class modelling using curve matching. In: IEEE Conference on Computer
  Vision and Pattern Recognition

\bibitem[{Ramakrishna et~al(2012)Ramakrishna, Kanade, and
  Sheikh}]{ramakrishna2012reconstructing}
Ramakrishna V, Kanade T, Sheikh Y (2012) Reconstructing 3d human pose from 2d
  image landmarks. In: European Conference on Computer Vision

\bibitem[{Ren et~al(2015)Ren, He, Girshick, and Sun}]{FastRCNN}
Ren S, He K, Girshick R, Sun J (2015) Faster {R-CNN}: Towards real-time object
  detection with region proposal networks. In: Advances in Neural Information
  Processing Systems

\bibitem[{Russakovsky et~al(2015)Russakovsky, Deng, Su, Krause, Satheesh, Ma,
  Huang, Karpathy, Khosla, Bernstein et~al}]{russakovsky2015imagenet}
Russakovsky O, Deng J, Su H, Krause J, Satheesh S, Ma S, Huang Z, Karpathy A,
  Khosla A, Bernstein M, et~al (2015) Imagenet large scale visual recognition
  challenge. International Journal of Computer Vision 115(3):211--252

\bibitem[{Sapp and Taskar(2013)}]{FLIC}
Sapp B, Taskar B (2013) Modec: Multimodal decomposable models for human pose
  estimation. In: IEEE Conference on Computer Vision and Pattern Recognition

\bibitem[{Satkin et~al(2012)Satkin, Lin, and Hebert}]{satkin_bmvc2012}
Satkin S, Lin J, Hebert M (2012) Data-driven scene understanding from 3{D}
  models. In: British Machine Vision Conference

\bibitem[{Shakhnarovich et~al(2003)Shakhnarovich, Viola, and
  Darrell}]{shakhnarovich2003fast}
Shakhnarovich G, Viola P, Darrell T (2003) Fast pose estimation with
  parameter-sensitive hashing. In: IEEE International Conference on Computer
  Vision

\bibitem[{Shih et~al(2015)Shih, Mallya, Singh, and Hoiem}]{shih2015part}
Shih KJ, Mallya A, Singh S, Hoiem D (2015) Part localization using
  multi-proposal consensus for fine-grained categorization. In: British Machine
  Vision Conference

\bibitem[{Shrivastava and Gupta(2013)}]{shrivastava2013building}
Shrivastava A, Gupta A (2013) Building part-based object detectors via 3d
  geometry. In: IEEE International Conference on Computer Vision

\bibitem[{Soltani et~al(2017)Soltani, Huang, Wu, Kulkarni, and
  Tenenbaum}]{Soltani2017Synthesizing}
Soltani AA, Huang H, Wu J, Kulkarni TD, Tenenbaum JB (2017) Synthesizing 3d
  shapes via modeling multi-view depth maps and silhouettes with deep
  generative networks. In: IEEE Conference on Computer Vision and Pattern
  Recognition

\bibitem[{Song et~al(2017)Song, Yu, Zeng, Chang, Savva, and
  Funkhouser}]{song2016semantic}
Song S, Yu F, Zeng A, Chang AX, Savva M, Funkhouser T (2017) Semantic scene
  completion from a single depth image. In: IEEE Conference on Computer Vision
  and Pattern Recognition

\bibitem[{Su et~al(2014)Su, Huang, Mitra, Li, and Guibas}]{haosu_sig14}
Su H, Huang Q, Mitra NJ, Li Y, Guibas L (2014) Estimating image depth using
  shape collections. ACM Transactions on Graphics 33(4):37

\bibitem[{Su et~al(2015)Su, Qi, Li, and Guibas}]{su15}
Su H, Qi CR, Li Y, Guibas L (2015) Render for cnn: Viewpoint estimation in
  images using cnns trained with rendered 3d model views. In: IEEE
  International Conference on Computer Vision

\bibitem[{Sun and Saenko(2014)}]{sun2014virtual}
Sun B, Saenko K (2014) From virtual to reality: Fast adaptation of virtual
  object detectors to real domains. In: British Machine Vision Conference

\bibitem[{Taigman et~al(2015)Taigman, Yang, Ranzato, and Wolf}]{deepface2}
Taigman Y, Yang M, Ranzato M, Wolf L (2015) Web-scale training for face
  identification. In: IEEE Conference on Computer Vision and Pattern
  Recognition

\bibitem[{Tompson et~al(2015)Tompson, Goroshin, Jain, LeCun, and
  Bregler}]{tompson2015efficient}
Tompson J, Goroshin R, Jain A, LeCun Y, Bregler C (2015) Efficient object
  localization using convolutional networks. In: IEEE Conference on Computer
  Vision and Pattern Recognition

\bibitem[{Tompson et~al(2014)Tompson, Jain, LeCun, and
  Bregler}]{tompson2014joint}
Tompson JJ, Jain A, LeCun Y, Bregler C (2014) Joint training of a convolutional
  network and a graphical model for human pose estimation. In: Advances in
  Neural Information Processing Systems

\bibitem[{Torralba and Efros(2011)}]{datasetbias}
Torralba A, Efros AA (2011) Unbiased look at dataset bias. In: IEEE Conference
  on Computer Vision and Pattern Recognition

\bibitem[{Torresani et~al(2003)Torresani, Hertzmann, and Bregler}]{non_rigid3d}
Torresani L, Hertzmann A, Bregler C (2003) Learning non-rigid 3d shape from 2d
  motion. In: Advances in Neural Information Processing Systems

\bibitem[{Toshev and Szegedy(2014)}]{toshev2014deeppose}
Toshev A, Szegedy C (2014) Deeppose: Human pose estimation via deep neural
  networks. In: IEEE Conference on Computer Vision and Pattern Recognition, pp
  1653--1660

\bibitem[{Tulsiani and Malik(2015)}]{tulsiani2015viewpoints}
Tulsiani S, Malik J (2015) Viewpoints and keypoints. In: IEEE Conference on
  Computer Vision and Pattern Recognition

\bibitem[{Tulsiani et~al(2017)Tulsiani, Zhou, Efros, and
  Malik}]{tulsiani2017multi}
Tulsiani S, Zhou T, Efros AA, Malik J (2017) Multi-view supervision for
  single-view reconstruction via differentiable ray consistency. In: IEEE
  Conference on Computer Vision and Pattern Recognition

\bibitem[{Vicente et~al(2014)Vicente, Carreira, Agapito, and
  Batista}]{vicente2014reconstructing}
Vicente S, Carreira J, Agapito L, Batista J (2014) Reconstructing pascal voc.
  In: IEEE Conference on Computer Vision and Pattern Recognition

\bibitem[{Wah et~al(2011)Wah, Branson, Welinder, Perona, and Belongie}]{CUB}
Wah C, Branson S, Welinder P, Perona P, Belongie S (2011) {The Caltech-UCSD
  Birds-200-2011 Dataset}. Tech. Rep. CNS-TR-2011-001, California Institute of
  Technology

\bibitem[{Wu et~al(2015)Wu, Yildirim, Lim, Freeman, and
  Tenenbaum}]{wu2015galileo}
Wu J, Yildirim I, Lim JJ, Freeman B, Tenenbaum J (2015) Galileo: Perceiving
  physical object properties by integrating a physics engine with deep
  learning. In: Advances in Neural Information Processing Systems

\bibitem[{Wu et~al(2016)Wu, Zhang, Xue, Freeman, and Tenenbaum}]{3dgan}
Wu J, Zhang C, Xue T, Freeman WT, Tenenbaum JB (2016) Learning a probabilistic
  latent space of object shapes via 3d generative-adversarial modeling. In:
  Advances in Neural Information Processing Systems

\bibitem[{Wu et~al(2017)Wu, Wang, Xue, Sun, Freeman, and Tenenbaum}]{marrnet}
Wu J, Wang Y, Xue T, Sun X, Freeman WT, Tenenbaum JB (2017) Marrnet: 3d shape
  reconstruction via 2.5d sketches. In: Advances in Neural Information
  Processing Systems

\bibitem[{Xiang et~al(2014)Xiang, Mottaghi, and Savarese}]{pascal3d}
Xiang Y, Mottaghi R, Savarese S (2014) Beyond pascal: A benchmark for 3d object
  detection in the wild. In: IEEE Winter Conference on Applications of Computer
  Vision

\bibitem[{Xiao et~al(2010)Xiao, Hays, Ehinger, Oliva, and Torralba}]{SUN}
Xiao J, Hays J, Ehinger K, Oliva A, Torralba A (2010) Sun database: Large-scale
  scene recognition from abbey to zoo. In: IEEE Conference on Computer Vision
  and Pattern Recognition

\bibitem[{Xue et~al(2012)Xue, Liu, and Tang}]{xue2012example}
Xue T, Liu J, Tang X (2012) Example-based 3d object reconstruction from line
  drawings. In: IEEE Conference on Computer Vision and Pattern Recognition

\bibitem[{Yang and Ramanan(2011)}]{yang2011articulated}
Yang Y, Ramanan D (2011) Articulated pose estimation with flexible
  mixtures-of-parts. In: IEEE Conference on Computer Vision and Pattern
  Recognition

\bibitem[{Yasin et~al(2016)Yasin, Iqbal, Kr{\"u}ger, Weber, and
  Gall}]{yasin2016dualsource}
Yasin H, Iqbal U, Kr{\"u}ger B, Weber A, Gall J (2016) A dual-source approach
  for 3d pose estimation from a single image. In: IEEE Conference on Computer
  Vision and Pattern Recognition

\bibitem[{Yuille and Kersten(2006)}]{yuille2006vision}
Yuille A, Kersten D (2006) Vision as bayesian inference: analysis by synthesis?
  Trends in cognitive sciences 10(7):301--308

\bibitem[{Zeng et~al(2016)Zeng, Song, Nie{\ss}ner, Fisher, and
  Xiao}]{zeng20163dmatch}
Zeng A, Song S, Nie{\ss}ner M, Fisher M, Xiao J (2016) 3dmatch: Learning the
  matching of local 3d geometry in range scans. In: IEEE Conference on Computer
  Vision and Pattern Recognition

\bibitem[{Zhou et~al(2016)Zhou, Kr{\"a}henb{\"u}hl, Aubry, Huang, and
  Efros}]{zhou2016learning}
Zhou T, Kr{\"a}henb{\"u}hl P, Aubry M, Huang Q, Efros AA (2016) Learning dense
  correspondence via 3d-guided cycle consistency. In: IEEE Conference on
  Computer Vision and Pattern Recognition

\bibitem[{Zhou et~al(2015)Zhou, Leonardos, Hu, and Daniilidis}]{zhou153d}
Zhou X, Leonardos S, Hu X, Daniilidis K (2015) 3d shape reconstruction from 2d
  landmarks: A convex formulation. In: IEEE Conference on Computer Vision and
  Pattern Recognition

\bibitem[{Zia et~al(2013)Zia, Stark, Schiele, and Schindler}]{zia2013detailed}
Zia MZ, Stark M, Schiele B, Schindler K (2013) Detailed 3d representations for
  object recognition and modeling. IEEE Transactions on Pattern Analysis and
  Machine intelligence 35(11):2608--2623

\end{thebibliography}

\end{document}